\documentclass{article}
\usepackage{amsmath}
\usepackage[preprint]{neurips_2023}

\usepackage[utf8]{inputenc} % allow utf-8 input
\usepackage[T1]{fontenc}    % use 8-bit T1 fonts
\usepackage{hyperref}       % hyperlinks
\usepackage{url}            % simple URL typesetting
\usepackage{booktabs}       % professional-quality tables
\usepackage{amsfonts}       % blackboard math symbols
\usepackage{nicefrac}       % compact symbols for 1/2, etc.
\usepackage{abstract}       % better abstract handling
\usepackage{microtype}      % microtypography
\usepackage{xcolor}         % colors

\usepackage{eucal}
\usepackage{amsmath, amssymb, amsthm}
\usepackage{bm}
\usepackage{graphicx}
\usepackage{float}
\usepackage{subcaption}
\usepackage{multirow}
\usepackage[ruled, vlined]{algorithm2e}
\usepackage{tikz}
\usepackage{pgfplots}
\usepackage{algorithm2e}
\usepackage{algorithmic}
\usepackage{booktabs}
\usepackage{array}
\usepackage{xr}
\usetikzlibrary{arrows.meta, positioning}
\pgfplotsset{compat=1.8}
\usepgfplotslibrary{groupplots}
\usepgfplotslibrary{fillbetween}

\definecolor{color0}{HTML}{636EFA}
\definecolor{color1}{HTML}{EF553B}
\definecolor{color2}{HTML}{00CC96}
\definecolor{color3}{HTML}{AB63FA}
\definecolor{color4}{HTML}{FFA15A}
\definecolor{color5}{HTML}{19D3F3}
\definecolor{color6}{HTML}{FF6692}
\definecolor{color7}{HTML}{B6E880}
\definecolor{color8}{HTML}{FF97FF}
\definecolor{color9}{HTML}{FECB52}

\newcommand{\ind}{{\;\perp\!\!\!\perp\;}}

\newtheorem{theorem}{Theorem}

\title{Scalable Causal Discovery from Recursive Nonlinear Data via Truncated Basis Function Scores and Tests}

\author{
  Joseph Ramsey \\
  Department of Philosophy \\
  Carnegie Mellon University \\
  Pittsburgh, PA 15213 \\
  \texttt{jdramsey@andrew.cmu.edu } \\
  \And
   Bryan Andrews \\
  Department of Psychiatry \& Behavioral Sciences \\
  University of Minnesota \\
  Minneapolis, MN 55454 \\
  \texttt{andr1017@umn.edu} \\
  \And  
  Peter Spirtes \\
  Department of Philosophy \\
  Carnegie Mellon University \\
  Pittsburgh, PA 15213 \\ 
  \texttt{ps7z@andrew.cmu.edu }
}

\begin{document}

\maketitle

\begin{abstract}
Learning graphical conditional independence structures from nonlinear, continuous or mixed data is a central challenge in machine learning and the sciences, and many existing methods struggle to scale to thousands of samples or hundreds of variables.

We introduce two basis–expansion tools for scalable causal discovery. First, the \emph{Basis Function BIC} (BF-BIC) score uses truncated additive expansions to approximate nonlinear dependencies. BF-BIC is theoretically consistent under additive models and extends to post–nonlinear (PNL) models via an invertible reparameterization. It remains robust under moderate interactions and supports mixed data through a degenerate-Gaussian embedding for discrete variables. In simulations with fully nonlinear neural causal models (NCMs), BF-BIC outperforms kernel- and constraint-based methods (e.g., KCI, RFCI) in both accuracy and runtime.

Second, the \emph{Basis Function Likelihood Ratio Test} (BF-LRT) provides an approximate conditional independence test that is substantially faster than kernel tests while retaining competitive accuracy.

Extensive simulations and a real-data application to Canadian wildfire risk show that, when integrated into hybrid searches, BF-based methods enable interpretable and scalable causal discovery. Implementations are available in Python, R, and Java.
\end{abstract}

\section{Introduction}
\label{sec:introduction}

Causal discovery from nonlinear and mixed (continuous + discrete) data is a central challenge in modern statistics and machine learning. Existing approaches, such as kernel conditional independence tests or nonparametric regression models, often face scalability limits when applied to large sample sizes or high-dimensional graphs.

We propose two basis–expansion tools for recursive nonlinear Structural Equation Models (rSEMs): the \emph{Basis Function BIC} (BF-BIC) and the \emph{Basis Function Likelihood Ratio Test} (BF-LRT). Both use truncated orthogonal polynomial expansions of continuous variables to represent additive nonlinear dependencies in finite-dimensional linear spaces. BF-BIC is a scoring criterion for score-based search, while BF-LRT serves as an approximate conditional independence test for constraint-based search. For methods such as FCIT \citep{ramsey2025efficientlatentvariablecausal} that use both a score and a test, the two methods may be used in tandem.

BF-BIC generalizes the Degenerate Gaussian BIC (DG-BIC) of \citet{andrews2019learning} by embedding continuous variables in Legendre polynomial bases, while retaining DG-style embeddings for categorical variables. It integrates naturally with efficient algorithms such as Best-Order Score Search (BOSS; \citet{andrews2023fast}), enabling accurate CPDAG estimation at scale. BF-LRT applies the same basis framework to likelihood ratio testing, producing asymptotically $\chi^2$ p-values at far lower computational cost than kernel methods.

Although the working assumption is additive nonlinearity (each variable modeled as a sum of nonlinear functions of its parents), the methods generalize formally to the post–nonlinear (PNL) setting via an invertible reparameterization (Section~\ref{sec:post_nonlin} of the appendix). Thus, the additive formulation is not restrictive in practice.

As a real-data demonstration, we apply BF-BIC and BF-LRT within the FCIT algorithm to the Algerian Forest Fire dataset. This hybrid search recovers plausible nonlinear causal structure, including known relationships such as $ \text{FWI} \rightarrow \text{Fire}$, while supporting latent-variable discovery via Partial Ancestral Graphs (PAGs). We also note a recent astrophysics note \citep{desmond2025causalstructuregalacticastrophysics} using FCIT \citep{ramsey2025efficientlatentvariablecausal} with both BF-BIC and BF-LRT to model galaxy data. These applications illustrate the compatibility of our methods with latent-variable causal discovery and their utility for scientific modeling in domains with both deterministic indices and hidden confounders. 

The remainder of the paper is organized as follows. Section~\ref{bf_tools_section} introduces BF-BIC and BF-LRT formally. Section~\ref{using_bf_methods_section} explains their integration into causal search algorithms. Section~\ref{simulation} reports extensive simulation results, and Section~\ref{real_data_example} presents the wildfire application. Section~\ref{sec:discussion} summarizes findings, and Section~\ref{conclusion} outlines future directions.

Our primary contributions are:
\begin{enumerate}
    \item A novel BF-BIC score enabling scalable nonlinear CPDAG estimation with efficient algorithms such as BOSS, theoretically consistent under additive models and empirically robust beyond them.
    \item A novel BF-LRT providing fast approximate conditional independence testing for constraint-based frameworks such as PC and PC-Max.
    \item Extensive simulations validating the effectiveness of both tools in small- and large-scale scenarios, including fully nonlinear neural data-generating processes.
    \item Rigorous theoretical justification of consistency under exponential-family noise and sufficient basis expansion, with extension to post-nonlinear models.
    \item A real-data application combining BF-BIC and BF-LRT with FCIT for latent-variable causal discovery, recovering interpretable nonlinear PAGs.
\end{enumerate}

\section{Related Work}
\label{sec:related_work}

Causal discovery in nonlinear settings has developed along several main lines. A first class builds on \emph{additive noise models (ANMs)}, where each variable is a nonlinear function of its parents plus independent noise. Under mild assumptions, such models are identifiable from observational data \citep{hoyer2009nonlinear, peters2014causal}. The \emph{Causal Additive Model (CAM)} \citep{buhlmann2014cam} exemplifies this approach, combining greedy order search with penalized regression. CAM guarantees recovery under smoothness and sparsity assumptions, but focuses on variable ordering and continuous data. In contrast, our approach represents nonlinearities with truncated orthogonal basis expansions and applies score- or test-based procedures directly, supporting scalable algorithms such as BOSS \citep{andrews2023fast} or PC-Max, and handling mixed data through DG embeddings.

A second line extends \emph{score-based search} to nonlinear settings, using kernel scores, generalized additive models, or spline regression \citep{peters2014causal, zhang2020nonlinear, huang2020generalized}. These methods are flexible but computationally costly due to repeated nonparametric estimation. Our Basis Function BIC (BF-BIC) sidesteps kernel smoothing by embedding nonlinearities into fixed linear spaces, yielding faster optimization without loss of accuracy.

In the \emph{constraint-based setting}, nonlinear conditional independence tests have relied on reproducing kernel Hilbert space (RKHS) methods such as KCI \citep{zhang2012kernel} and its randomized variant RCIT \citep{strobl2019approximate}. While powerful, these tests are expensive for large $n$ or conditioning sets and lack a clear model-based consistency guarantee. Our Basis Function Likelihood Ratio Test (BF-LRT) instead applies generalized likelihood ratio testing to basis-expanded variables, providing asymptotic $\chi^2$ guarantees under additive assumptions while remaining lightweight in practice.

A third thread explores \emph{deep generative models}, including post-nonlinear ICA \citep{zhang2009identifiability,zhang2010nonlinear} and VAE-based causal models \citep{louizos2017causal}. These offer great flexibility but often at the expense of interpretability and scalability. Our framework is simpler and more interpretable, while still accommodating flexible nonlinearities through basis expansions.

Finally, work on the \emph{post-nonlinear (PNL)} framework shows that invertible transformations preserve identifiability advantages of ANMs \citep{zhang2009identifiability,zhang2010nonlinear}. Our methods extend naturally to this setting: once an appropriate transform $g^{-1}$ is applied, BF-BIC and BF-LRT operate within the same additive-basis framework (see Section~\ref{sec:post_nonlin}). This connection ties modern causal discovery to classical approximation theory, where truncated orthogonal expansions (e.g., Legendre polynomials) have long been used for regression and density estimation \citep{newey1997convergence}. Leveraging orthogonality yields numerically stable, scalable procedures that remain effective even beyond strict additivity.

In summary, our work bridges the gap between flexible but costly kernel methods, additive approaches like CAM, and deep generative models. By embedding nonlinearities in orthogonal bases, we provide interpretable, scalable, and theoretically grounded tools compatible with both score-based and constraint-based causal discovery.

\section{Preliminaries}

We begin by defining key concepts from graphical causal models, focusing on models based on \emph{directed acyclic graphs (DAGs)} and their extensions.

\paragraph{DAGs.}
A DAG consists of nodes (variables) and directed edges \(X \rightarrow Y\), where each directed edge represents that variable \(X\) \emph{causes} variable \(Y\). Here, causation means that interventions on \(X\) counterfactually alter \(Y\), at least for some values \citep{spirtes2001causation}.  

We use non-boldface capital letters to denote individual variables and boldface letters to denote sets of distinct variables. A \emph{directed graph} \(G\) over a set of variables \(\mathbf{V}\) consists of directed edges whose endpoints are in \(\mathbf{V}\). Each edge \(X \rightarrow Y\) has a \emph{tail} at \(X\) and an \emph{arrowhead} at \(Y\).  

A \emph{path} \(P\) in \(G\) from \(X_1\) to \(X_n\) is a sequence of distinct variables \(\langle X_1, X_2, \dots, X_n \rangle\), \(n \geq 1\), such that each consecutive pair \((X_i, X_{i+1})\) is joined by either \(X_i \rightarrow X_{i+1}\) or \(X_i \leftarrow X_{i+1}\). A path is \emph{cyclic} if it contains the same variable more than once; otherwise it is \emph{acyclic}. A graph is a DAG if and only if all paths are acyclic.  

Two variables \(X\) and \(Y\) are \emph{adjacent} in \(G\) if there is an edge between them. A \emph{collider} along a path is a subsequence \(\langle A, B, C \rangle\) with \(A \rightarrow B \leftarrow C\). If \(A\) and \(C\) are not adjacent in \(G\), this is called an \emph{unshielded collider}. A variable \(X\) is an \emph{ancestor} of \(Y\) if there exists a directed path from \(X\) to \(Y\) (equivalently, \(Y\) is a \emph{descendant} of \(X\)). If there is a directed edge \(W \rightarrow X\), then \(W\) is a \emph{parent} of \(X\).  

\paragraph{d-separation.}
DAGs encode conditional independence constraints via \emph{d-separation}. Let \(X, Y\) be distinct variables and \(\mathbf{O} \subseteq \mathbf{V}\setminus\{X,Y\}\). A path between \(X\) and \(Y\) is \emph{d-connected} given \(\mathbf{O}\) if and only if  
(1) every collider on the path has a descendant in \(\mathbf{O}\), and  
(2) no non-collider on the path is in \(\mathbf{O}\).  
Otherwise the path is \emph{d-separated}. Two sets of variables \(\mathbf{X}\) and \(\mathbf{Y}\) are d-connected given \(\mathbf{O}\) if there exists at least one d-connected path between them; otherwise they are d-separated given \(\mathbf{O}\).

\paragraph{CPDAGs.}
A \emph{Completed Partially Directed Acyclic Graph (CPDAG)} extends DAGs by allowing \emph{undirected edges} \(X - Y\), where both endpoints have tails. An undirected edge indicates uncertainty about the direction: either \(X \rightarrow Y\) or \(X \leftarrow Y\) could hold, but the choice does not affect the implied conditional independencies. A \emph{Markov equivalence class} of DAGs is a set of DAGs implying exactly the same conditional independencies. A CPDAG is the graphical representation of such a class: all unshielded colliders are oriented as in the CPDAG, and all remaining undirected edges may be oriented arbitrarily without creating new unshielded colliders \citep{spirtes2001causation}.

\paragraph{Recursive SEMs.}
A \emph{recursive structural equation model (rSEM)} over a DAG \(G=(\mathbf{V},E)\) defines a statistical model in which each \(X \in \mathbf{V}\) is expressed as
\[
X = f(W_1, \dots, W_n, e_X),
\]
where \(W_1,\dots,W_n\) are the parents of \(X\) and \(e_X\) is an independent \emph{exogenous noise variable}. An rSEM is \emph{linear} if each function \(f\) is linear; otherwise it is \emph{nonlinear}.  

\emph{Additive nonlinear models}, a special case, assume
\[
X = f_1(W_1) + f_2(W_2) + \dots + f_n(W_n) + e_X,
\]
excluding interaction terms (e.g., \(W_1W_2\)). This representation simplifies inference while remaining expressive in many settings.

\paragraph{Basis expansions.}
In our analysis, we expand variables into orthogonal polynomial bases (e.g., Legendre polynomials). This produces explicit nonlinear features of the original variables. Importantly, exponential-family distributions of the exogenous noise terms remain exponential-family under linear transformations of these expansions \citep[Chapter~1]{brown1986fundamentals}, a fact we leverage in our theoretical justification.

\paragraph{MAGs and m-separation.}
A \emph{Maximal Ancestral Graph (MAG)} generalizes DAGs to incorporate latent variables and selection bias. MAGs are mixed graphs containing directed (\(\rightarrow\)) and bidirected (\(\leftrightarrow\)) edges. Directed edges denote direct causal relationships among observed variables, while bidirected edges represent unobserved common causes. Every missing edge corresponds to a conditional independence, and no additional edge can be added without violating this property—hence “maximal.”  

MAGs generalize d-separation into \emph{m-separation}, which accounts for latent confounding and selection bias.

\paragraph{PAGs.}
A \emph{Partial Ancestral Graph (PAG)} represents a Markov equivalence class of MAGs, encoding all conditional independencies that hold in every compatible MAG. PAGs are partially oriented and allow a richer set of edge types:
\begin{itemize}
    \item \(X \rightarrow Y\): compelled directed edge; \(X\) is an ancestor of \(Y\) in all MAGs.
    \item \(X \leftrightarrow Y\): bidirected edge, representing a latent confounder.
    \item \(X \circ\!\!\!\!\rightarrow Y\): circle–arrow edge; \(Y\) is not an ancestor of \(X\), but orientation at \(X\) is unresolved.
    \item \(X \circ\!-\!\circ Y\): circle–circle edge; adjacency is known, but neither endpoint orientation is compelled.
    \item \(X - Y\): tail–tail edge, used in CPDAGs and occasionally appearing in PAG representations.
\end{itemize}
Here the circle endpoint \(\circ\) denotes uncertainty about whether that endpoint is a tail or arrowhead.  

\paragraph{Our setting.}
In our work we allow for latent variables by applying the FCIT algorithm, which searches over PAGs using a hybrid test-and-score strategy. Combined with basis function methods, FCIT can recover nonlinear causal relationships while accounting for hidden confounders and selection bias.

\section{Basis Function BIC and Basis Function LRT}
\label{bf_tools_section}

The Basis Function Bayesian Information Criterion (BF-BIC) modifies the existing Degenerate Gaussian Bayesian Information Criterion (DG-BIC). DG-BIC extends linear BIC scoring to accommodate multinomial variables by embedding categorical indicators. For a multinomial variable with \( c \) categories, DG-BIC introduces indicators for \( c-1 \) categories to circumvent singularity in score calculations. Detailed theoretical underpinnings of DG-BIC are provided in \citep{andrews2019learning}. Notably, DG-BIC presupposes linear relationships among continuous variables and does not perform further embedding on them. Additive structural models, in which each variable is modeled as the sum of functions over its individual parent variables, have been studied extensively in the context of causal discovery. Notable examples include Causal Additive Models (CAM) \citep{buhlmann2014cam}, which use penalized regression and high-dimensional order search, and Additive Noise Models (ANMs) \citep{hoyer2009nonlinear, peters2014causal}, which establish identifiability results under independence between noise and inputs. Our BF-BIC method assumes a similar additive structure, but represents the component functions using truncated orthogonal basis expansions, enabling score-based causal discovery using efficient linear algebra operations and integration with high-performance search algorithms such as BOSS.

BF-BIC expands DG-BIC by additionally embedding continuous variables, thereby allowing for nonlinear but additive relationships. Each continuous variable is substituted by a set of non-constant columns generated via a basis expansion.\footnote{Basis expansions typically start with a constant term \( 1 \), followed by \( x \), to capture linear dependencies, then further higher-order terms. Here, we omit the constant term explicitly, although intercept terms are included in regressions.} This expansion may involve polynomial bases or orthogonal bases such as Legendre, Hermite, or Chebyshev polynomials, all of which offer efficient recursive formulations. We employ the Legendre basis, advantageous for numerical stability due to its confinement to the interval \([-1, 1]\), limiting the magnitude of higher-power terms. The basis expansion is truncated at \( p \) terms, where \( p \) is configurable by the user. For multinomial variables, we retain categorical indicator embedding from DG-BIC.

The BIC score over embedded variables is calculated as follows. Given embedded variables \( X \) and \( \mathbf{Z} \), each containing three columns:

\[
X = \langle X_1, X_2, X_3 \rangle, \quad \mathbf{Z} = \langle Z_1, Z_2, Z_3 \rangle,
\]

the conditional BIC \( BIC(X \mid \mathbf{Z}) \) is computed as:

\[
 BIC(X_1 \mid Z_1, Z_2, Z_3) + BIC(X_2 \mid Z_1, Z_2, Z_3, X_1) + BIC(X_3 \mid Z_1, Z_2, Z_3, X_1, X_2).
\]

This decomposition is valid when the basis captures the true structural function in an additive form. Each component score is obtained using a penalized likelihood:

\[
L = -\frac{N}{2} \log(2\pi \sigma^2) + 1,
\]

\[
BIC = 2L - c k \ln N,
\]

where \( \sigma^2 \) is the residual variance, \( k \) denotes the number of predictors in the linear model, \( N \) is the sample size, and \( c \) is a multiplier (``penalty discount'') on the penalty \citep{haughton1988choice}. The penalty discount adjusts the complexity penalty in BIC, allowing more flexibility for smaller sample sizes or more complex models. To improve computational efficiency, residual variances are computed using covariance matrices.

BF-BIC, like DG-BIC, is score-equivalent \citep{chickering2002optimal}, ensuring all Directed Acyclic Graphs (DAGs) within a Completed Partially Directed Acyclic Graph (CPDAG) equivalence class receive identical scores. As the truncation limit increases without bound, additive structural functions \( f(X) = f_1(Z_1) + f_2(Z_2) + \dots \) are represented arbitrarily well or exactly. Hence, when the structural function is additive and the truncation limit is sufficient, BF-BIC is theoretically consistent in the limit of large sample size. The theoretical justification and assumptions underpinning this convergence are detailed in Appendix~\ref{appendix:theory}.

In the general case—when the structural functions include interactions or other non-additive components—BF-BIC remains a flexible and scalable approximation. Although the residuals may then deviate from exponential-family form, the framework can be extended to the post-nonlinear setting by applying an inverse mapping to the additive nonlinear component of each structural equation.

BF-LRT employs the same embedding strategy as BF-BIC but utilizes a Generalized Likelihood Ratio Test (GLRT) for conditional independence testing, specifically evaluating:

\[
X \ind Y \mid \mathbf{Z}.
\]

First, we extract embedded columns for variables: \( X = \langle X_1, \dots, X_{x_{\max}} \rangle \), \( Y = \langle Y_1, \dots, Y_{y_{\max}} \rangle \), and \( \mathbf{Z} = \langle Z_1, \dots, Z_{z_{\max}} \rangle \). The embeddings of \( Y \) and \( \mathbf{Z} \) form the predictor matrix \( C_{yz} \).

We then calculate residual variances under two models:

\begin{itemize}
\item \textbf{Null model}: Regress \( X \) linearly onto \( \mathbf{Z} \) only, obtaining residual variance:
\[
\sigma_0^2 = \text{Var}(\epsilon_0), \quad \epsilon_0 = X - \hat{X}(Z).
\]

\item \textbf{Alternative model}: Regress \( X \) linearly onto \( Y \) and \( \mathbf{Z} \), obtaining residual variance:
\[
\sigma_1^2 = \text{Var}(\epsilon_1), \quad \epsilon_1 = X - \hat{X}(Y, Z).
\]
\end{itemize}

The likelihood ratio statistic is computed as:

\[
LR_{\text{stat}} = N \log\left( \frac{\sigma_0^2}{\sigma_1^2} \right).
\]

The associated p-value is derived using the chi-square cumulative distribution function (CDF):

\[
p = 1 - F_{\chi^2}(LR_{\text{stat}}, y_{\max}).
\]

The theoretical justification for BF-LRT is provided in Appendix~\ref{appendix:theory}. As with BF-BIC, the formal guarantees hold in the additive case under exponential-family assumptions, and extend to post-nonlinear models formed by applying invertible transformations to the additive components.

As mentioned, we are using truncations of the Legendre polynomial basis for our score. The method for calculating these values recursively is given in Algorithm~\ref{alg:legendre_recursive}. This is applied to variables in the data after they have been linearly scaled to the range \([-1, 1]\). The constant term, \(P_n(0)\), is not used. 

Other approaches to nonlinear causal discovery include general non-Gaussian SEMs \citep{zhang2010nonlinear} and deep generative models for causal discovery \citep{zhang2016causal}. These models typically emphasize functional identifiability under minimal assumptions, often at the cost of scalability. In contrast, our approach focuses on scalability through a truncated basis representation, offering a practical tradeoff between flexibility and interpretability.

\begin{algorithm}
\caption{Recursive Computation of Legendre Polynomials}
\label{alg:legendre_recursive}
\begin{algorithmic}[1]
\REQUIRE Integer $n \geq 0$, real number $x$
\ENSURE Value of Legendre polynomial $P_n(x)$
\IF{$n < 0$}
    \STATE \textbf{error} "Index must be non-negative"
\ELSIF{$n = 0$}
    \RETURN 1
\ELSIF{$n = 1$}
    \RETURN $x$
\ELSE
    \RETURN $((2n - 1) \cdot x \cdot \text{Legendre}(n - 1, x) - (n - 1) \cdot \text{Legendre}(n - 2, x)) / n$
\ENDIF
\end{algorithmic}
\end{algorithm}

\section{Using BF-BIC and BF-LRT for Inferring CPDAG Models}
\label{using_bf_methods_section}

BF-BIC and BF-LRT were developed to support causal search algorithms for estimating nonlinear CPDAG models—that is, Markov equivalence classes of DAGs without latent confounders. In practice, they can serve as plug-in tools wherever a nonlinear score or conditional independence test is required, or both.

BF-BIC is designed for score-based search methods such as FGES \citep{ramsey2017million}, BOSS \citep{andrews2023fast}, and GRaSP \citep{lam2022greedy}, while BF-LRT is suited to constraint-based methods such as PC \citep{spirtes2001causation} or PC-Max \citep{ramsey2016improving}. Because BOSS already demonstrates strong accuracy in the linear Gaussian case, we anticipate that this advantage extends to additive nonlinear models and, heuristically, to more general cases including multinomial variables.

\subsection{BF-BIC and Score-Based Search}

BF-BIC introduces two key user parameters:  
- the \emph{truncation limit} \(p\), controlling the number of basis terms for each continuous variable, and  
- the \emph{penalty discount} \(c\), scaling the BIC complexity penalty.  

These parameters allow a balance between flexibility, interpretability, and computational efficiency.

\paragraph{The BOSS algorithm.}
BOSS estimates CPDAGs by optimizing variable orderings:\footnote{In \citep{andrews2023fast}, an additional backward equivalence search (BES) step is discussed but found to be rarely impactful for large models and is omitted here.}
\begin{enumerate}
    \item \textbf{Initialization}: Begin with an initial permutation of variables and score the model.
    \item \textbf{Forward sweep}: Move the last variable into each position in the permutation, score each configuration, and keep the best-scoring model.
    \item \textbf{Iteration}: Repeat the sweep for each remaining variable.
    \item \textbf{Convergence}: Continue until no further improvements are found.
    \item \textbf{Output}: Return the CPDAG corresponding to the highest-scoring DAG.
\end{enumerate}

Because BF-BIC is score equivalent \citep{chickering2002optimal}, all DAGs in a CPDAG class share the same score. Thus, with additive structural functions adequately captured by the basis expansion, the optimal score corresponds to the correct CPDAG. For more general nonlinear functions, BF-BIC provides a scalable but approximate scoring criterion (see Appendix~\ref{appendix:theory}).

\subsection{BF-LRT and the PC-Max Algorithm}

BF-LRT uses a significance threshold \(\alpha\) to determine conditional independence, replacing the penalty discount of BF-BIC.

To mitigate collider orientation errors in traditional PC, we adopt PC-Max \citep{ramsey2016improving}, which orients unshielded triples using the separating set that maximizes the p-value. The steps are:

\begin{enumerate}
    \item \textbf{Initialization}: Start with a fully connected undirected graph.
    \item \textbf{Edge removal}: Iteratively remove edge \(X-Y\) if conditional independence is detected:
    \begin{itemize}
        \item \emph{Depth 0}: test unconditional independence \(X \ind Y\).
        \item \emph{Depth 1}: test independence \(X \ind Y \mid Z\) for each single neighbor \(Z\).
        \item \emph{Higher depths}: test on larger conditioning sets until no further edges can be removed.
    \end{itemize}
    \item \textbf{Collider orientation}: For each unshielded triple \(X-Y-Z\), orient as \(X \rightarrow Y \leftarrow Z\) if \(Y\) is not in the separating set that maximizes the p-value for \(X \ind Z \mid S\).
    \item \textbf{Propagation}: Orient additional edges to avoid new unshielded colliders and maintain consistency.
    \item \textbf{Output}: Return the resulting CPDAG.
\end{enumerate}

The main parameters are the truncation limit \(p\) and significance threshold \(\alpha\). BF-LRT is theoretically justified under additive exponential-family assumptions (Appendix~\ref{appendix:theory}), but also functions as an efficient and effective heuristic in broader nonlinear scenarios.

\section{Simulation Comparisons}
\label{simulation}

To generate simulated data for nonlinear rSEMs we use the random NCM described in Appendix Section \ref{appendix:cpn_section}.

We compare BOSS/BF-BIC, PC-Max/BF-LRT, and PC-Max/KCI \citep{zhang2012kernel} and PC-Max/RCIT \citep{strobl2019approximate} in smaller-scale simulations. We use KCI and RCIT from the causal-learn package \citep{zheng2024causal}. KCI uses kernel reasoning to ensure that $cov(f(X), g(Y)) = 0$ for all $f$ and $g$ for $X \ind Y$, for the unconditional case, and uses similar reasoning for the conditional case. Note that KCI in causal-learn is fully parallelized, with all matrix calculations optimized for multi-core execution. RCIT chooses random Fourier features to model non-linear relationships. Like BF-BIC, these methods consider functions of $X$ and $Y$, though in the context of the theorem that $cov(f(X), g(Y))$ being zero for all $f$ and $g$ if and only if $X \ind Y$. Fourier features are random sin or cosine functions of $X$ or $Y$. Because these implementations are slower than the BF methods for large $N$ and $p$, we include them only for the continuous 10 node average degree 2 cases.\footnote{All simulations are performed on a MacBook Pro M2 with 64 GB of RAM and 10 cores. RCIT scales fairly well though is dominated by the basis function methods. This may be partially due to its being implemented in Python, which is a slower language than Java. For speed, though, an implementation in C would be advantageous, which we have also pursued (https://github.com/bja43/causal-get).}

The CAM method proposed by \citet{buhlmann2014cam} is directly relevant to our setting, as it assumes a sparse, additive structural equation model with Gaussian noise.\footnote{For a feature comparison of CAM, additive noise models, and our BF-BIC and BF-LRT methods, please see Table~\ref{tab:additive_comparison} in the Appendix.} Although we had initially planned to include CAM in our benchmarking suite, we encountered technical challenges: the official R package for CAM has been removed from CRAN and is no longer actively maintained.  To provide a meaningful point of comparison, however, we replicated the experimental setup from Figure 3 of the CAM paper, generating random DAGs with 10 nodes, average in-degree 2, and 1000 samples, using additive nonlinear functions as in \citet{peters2014causal} with Gaussian noise. On this benchmark, BOSS with BF-BIC achieved an average SHD of 5.40 (±3.85), adjacency precision 1.00 (±0.00), and adjacency recall 0.90 (±0.06) over 10 runs. These results indicate strong performance in the same setting used to validate CAM.

\subsection{Simulation Setup}

We generate datasets under a range of structural and statistical conditions:

\begin{itemize}
    \item \textbf{Graph sizes:} 
    \begin{itemize}
        \item 10 nodes with 10 or 20 edges (average degrees 2 or 4).
        \item 20 nodes with 20 or 40 edges (average degrees 2 or 4).
    \end{itemize}
    
    \item \textbf{Sample sizes:} 
    200, 500, 1{,}000, 2{,}000, 5{,}000, and 10{,}000.
    
    \item \textbf{Data types:} 
    \begin{itemize}
        \item \emph{Continuous only:} Nonlinear functions generated by randomly initialized Neural Causal Models (NCMs; see Appendix~\ref{appendix:cpn_section}).
        \item \emph{Mixed:} A subset of variables rendered multinomial with 2–5 randomly assigned categories, with probability 0.2.
    \end{itemize}
    
    \item \textbf{Exogenous noise:} 
    Drawn independently from a $\mathrm{Beta}(2,5)$ distribution.
    
    \item \textbf{Repetitions:} 
    Ten independent dataset/DAG pairs per scenario.\footnote{Datasets are not stored in the GitHub repository, but random seeds are provided so results can be regenerated.}
    
    \item \textbf{Column order:} 
    Columns randomized to prevent order effects.
\end{itemize}

\subsection{CPDAG Estimation and Parameter Choices}

We estimated CPDAG models for each dataset using four methods: BOSS/BF-BIC, PC-Max/BF-LRT, PC-Max/KCI, and PC-Max/RCIT.  
BOSS, PC-Max, BF-BIC, and BF-LRT were implemented in \texttt{Tetrad}.  
KCI and RCIT were implemented using the \texttt{causal-learn} Python package.  

A scenario was included in the analysis if all searches for each of the 10 runs completed within 180 seconds.

\paragraph{Parameter tuning.}  
For BOSS/BF-BIC, the penalty discount parameter $c$ was tuned over $\{1, 2, 4, 8, 32, 64\}$ and truncation limits $p \in \{1, 3, 4, 8\}$.  
For PC-Max/BF-LRT, the same truncation limits were used, with significance thresholds $\alpha \in \{0.05, 0.01, 0.001\}$.  
For KCI and RCIT, we used default hyperparameters from the \texttt{causal-learn} package.  

Each estimated CPDAG was compared against the ground-truth CPDAG for accuracy evaluation.

\subsection{Evaluation Metrics}

For each method, the following metrics were computed and averaged over 10 runs:

\begin{itemize}
    \item \textbf{AP}: Adjacency Precision.  
    \item \textbf{AR}: Adjacency Recall.  
    \item \textbf{AHP}: Arrowhead Precision.  
    \item \textbf{AHR}: Arrowhead Recall.  
    \item \textbf{AHPC}: Arrowhead Precision restricted to edges in common with the generating model.  
    \item \textbf{AHRC}: Arrowhead Recall restricted to edges in common with the generating model.  
    \item \textbf{F1Adj}: F1 score for AP and AR.  
    \item \textbf{F1All}: Combined F1 score across AP, AR, AHP, and AHR.  
    \item \textbf{Elapsed}: Wall-clock runtime (seconds).  
\end{itemize}

We report these statistics for all scenarios in unified tables, indexed by algorithm, number of nodes, number of edges, sample size, truncation limit, penalty discount, and~$\alpha$ (the latter two where applicable).

\subsection{Results for Small-Scale Simulations}

\begin{figure}
    \centering
    \includegraphics[width=0.75\linewidth]{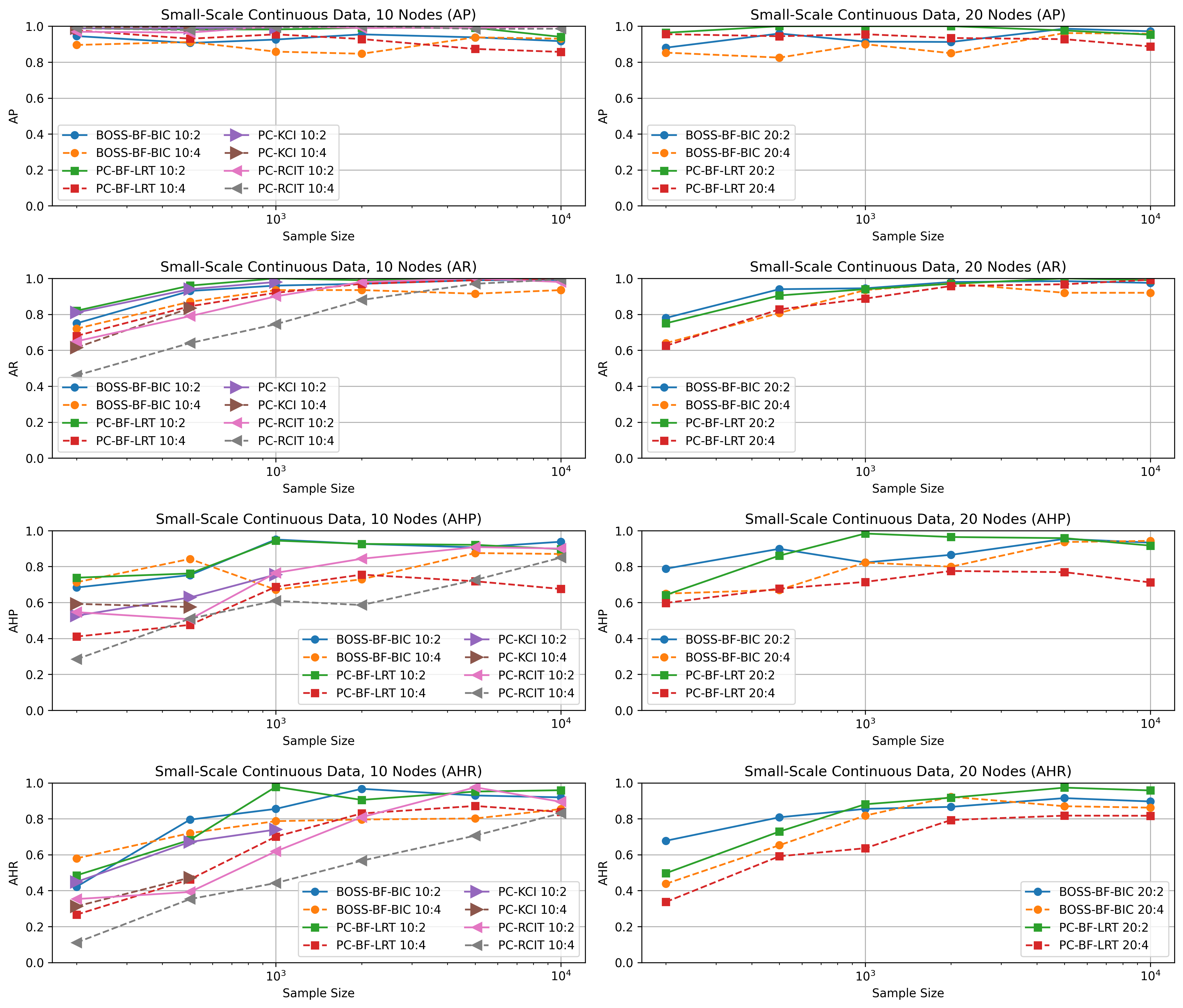}
    \caption{Evaluation plot for small scale continuous simulations. Each statistic plotted is an average over 10 runs and is the point selected by maximizing the F1Adj score.}
    \label{fig:continuous_small_scale}
\end{figure}

\begin{figure}
    \centering
    \includegraphics[width=0.75\linewidth]{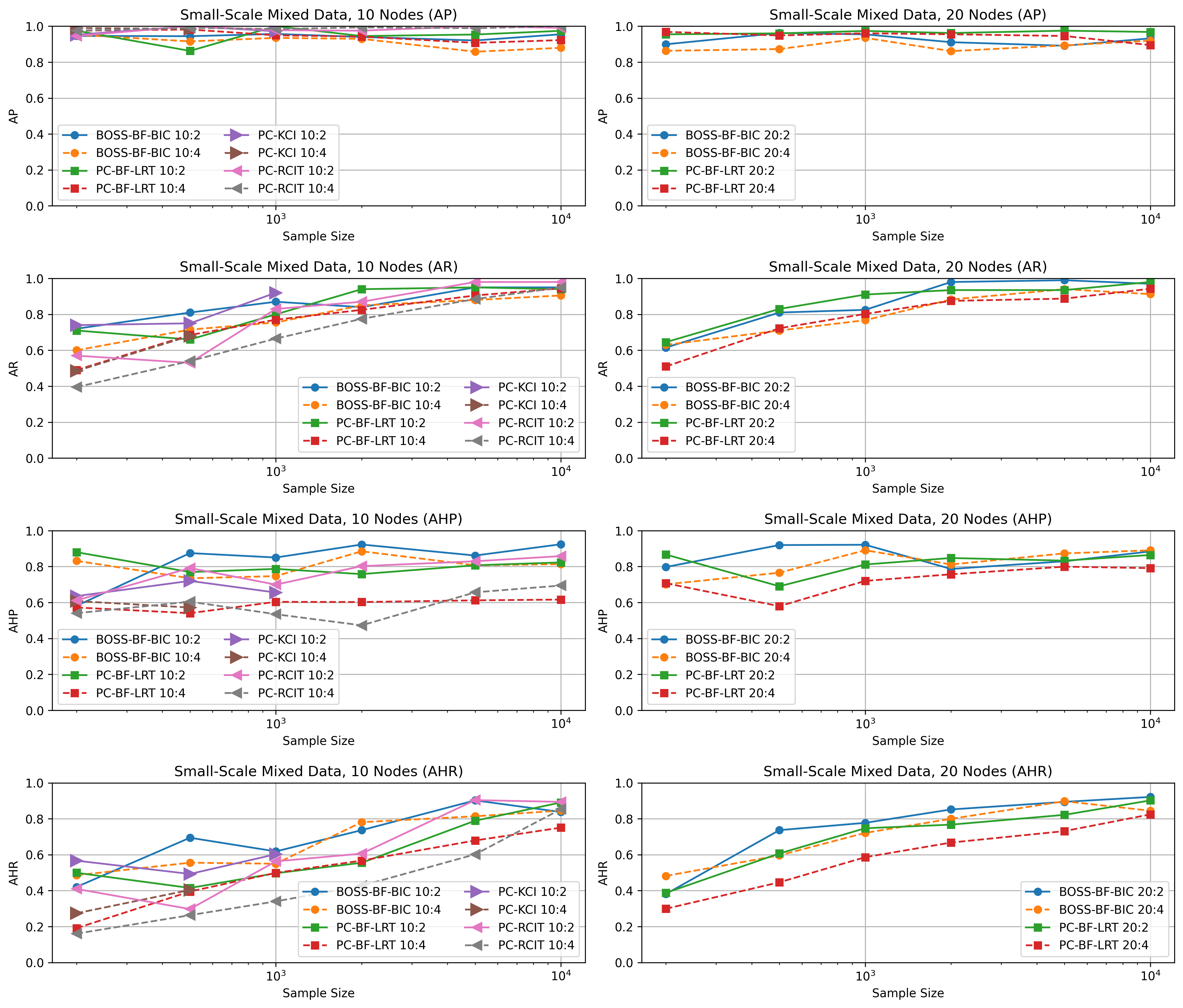}
    \caption{Evaluation plot for small scale mixed simulations. Each statistic plotted is an average over 10 runs and is the point selected by maximizing the F1Adj score.}
    \label{fig:mixed_small_scale}
\end{figure}

Figure \ref{fig:continuous_small_scale} presents AP, AR, AHP and AHR as a function of the sample size for the continuous case, selecting the model with the highest F1Adj score for each scenario. Figure \ref{fig:mixed_small_scale} shows similar results for the mixed case. The results are plotted for each algorithm and for the cases of continuous and mixed variables, for the following scenarios: 10 nodes, average degree 2,  10 nodes, average degree 4, 20 nodes, average degree 2, and 20 nodes, average degree 4. Only scenarios in which all runs are completed within 180 seconds are included. Full result tables are available on our GitHub site.

The results were generated as follows: Complete results were obtained for all scenarios on ten runs. For each combination of node count, edge count, and sample size, statistics were averaged over these ten runs for each algorithm. The rows were then sorted in descending order according to the F1Adj score and the rows with the highest score were recorded. Finally, AP, AR, AHP and AHR were plotted against the sample size for each algorithm and node/edge combination. The results are shown in Figures \ref{fig:continuous_small_scale} and \ref{fig:mixed_small_scale}. The timing results are given in Appendix Section~\ref{appendix:simulation_details}.

A discussion of results is given below, in Section \ref{sec:discussion}.

\subsection{Scaling Up: Large Sample Sizes and Large Graphs}

We also explored the performance of BOSS/BF-BIC and PC-Max/BF-LRT in settings with larger sample sizes. The simulation scenarios are as follows. (1) Scaling sample size $N$: 10 nodes with 10 or 20 edges, with sample sizes of 10,000, 20,000, 50,000, and 100,000. And (2) Scaling number of variables $p$: 50 nodes with 100 edges, or 100 nodes with 200 edges, with sample sizes of 500, 1000, 3000, 5000.

For the example in which the sample size is scaled, we used truncations 1, 3, 4, 5, 6, 7, and 8, penalty discounts 1, 2, 4, 8, 32, 64, 128, and alphas 0.1e-2, 0.1e-4, 0.1e-5, 1e-6, 1e-8, and 1e-10. For the example in which the number of variables is scaled, we used truncations 1, 3, and 4, penalty discounts 1, 2, 4, 6, and 8, and alphas 1e-2, 1e-4, 1e-5, and 1e-6. We did not include PC-Max/KCI or PC-Max/RCIT for the 20-variable smaller graphs or for the larger ranges, as they do not complete their searches within our time limit.
    
\begin{figure}
    \centering
    \includegraphics[width=0.75\linewidth]{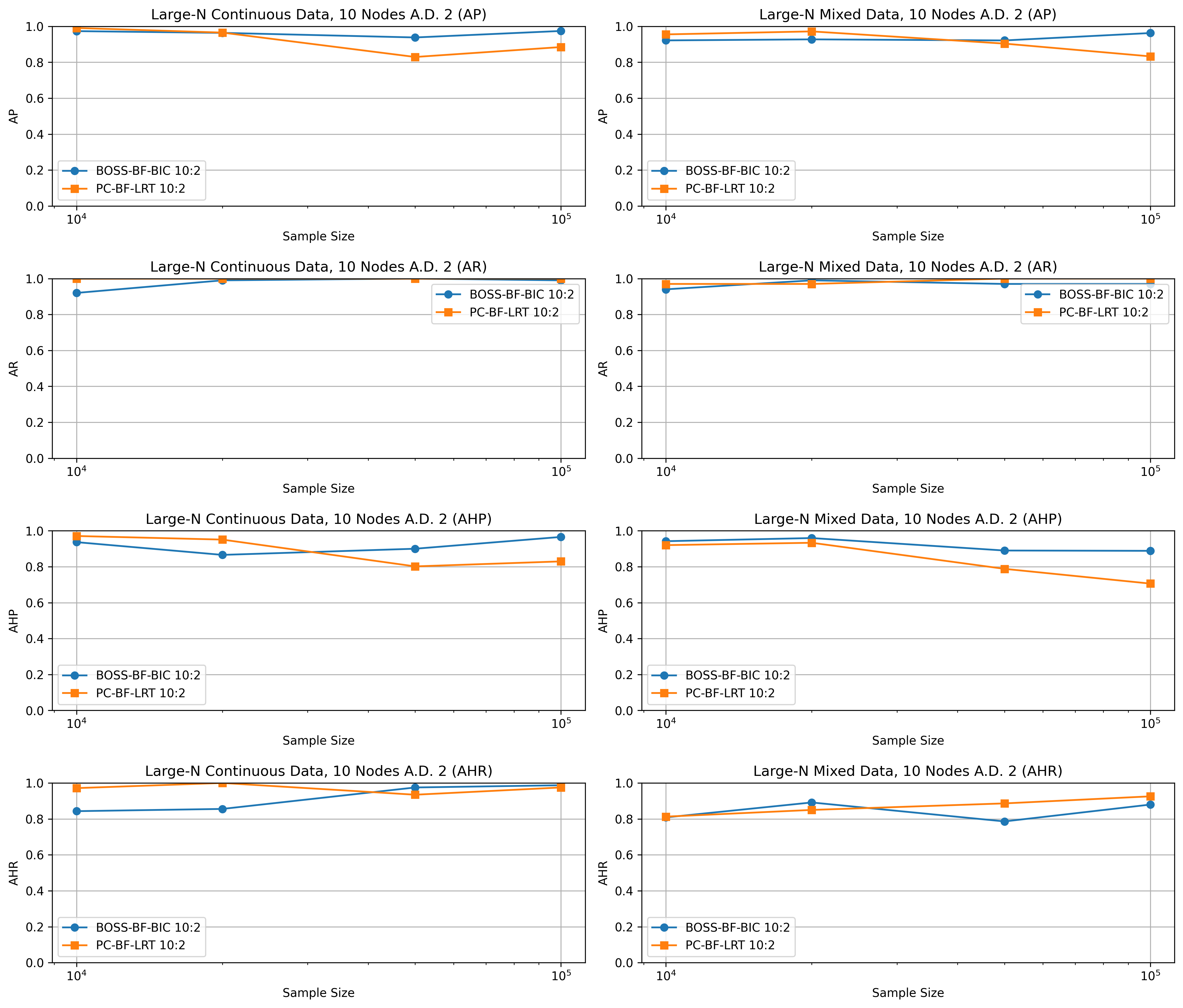}
    \caption{Evaluation Plot for large-N continuous simulations. Each statistic plotted is an average over 10 runs and is the point selected by maximizing the F1Adj score.}
    \label{fig:evaluation_large_n-continuous}
\end{figure}

\begin{figure}
    \centering
    \includegraphics[width=0.75\linewidth]{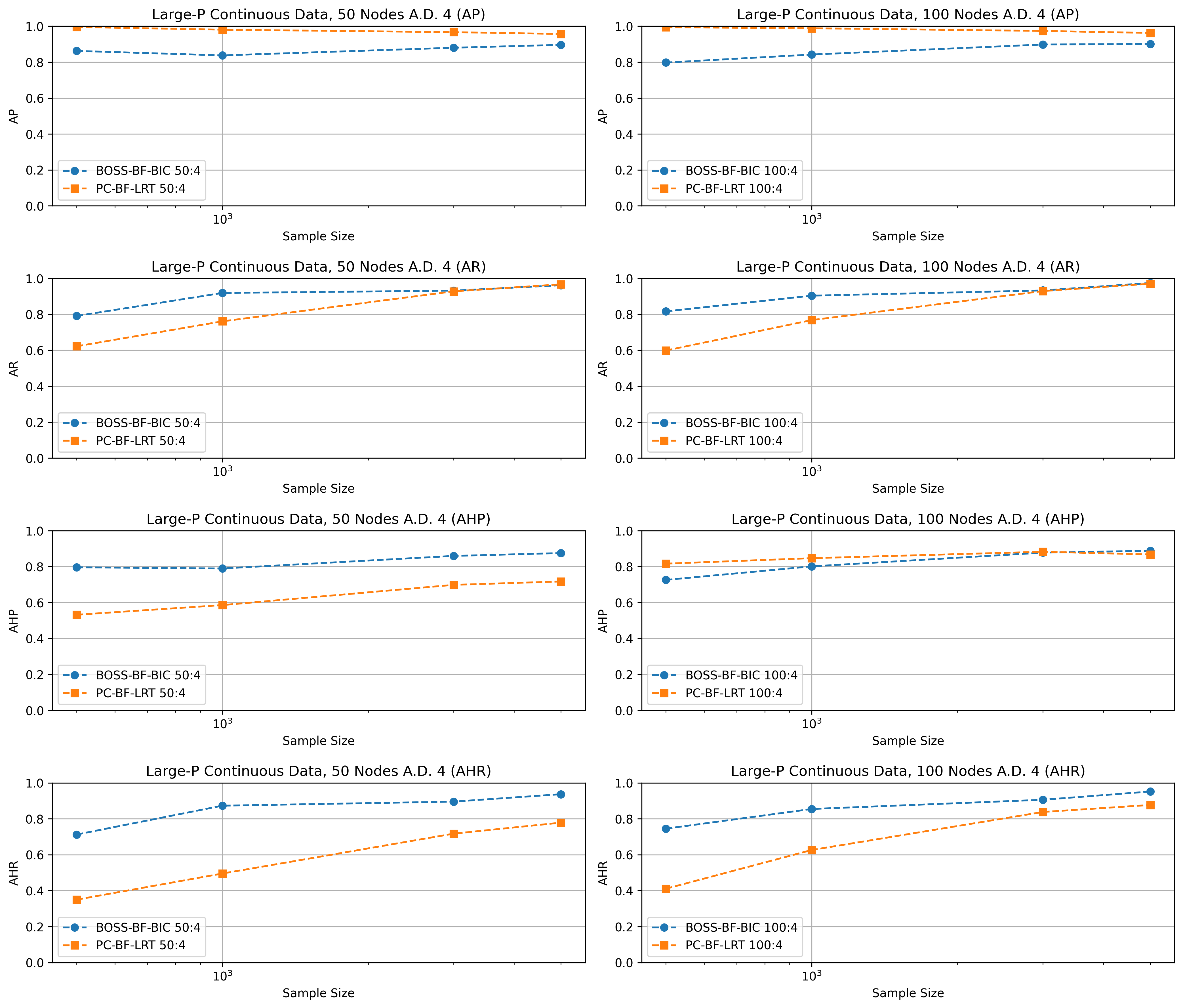}
    \caption{Evaluation Plot for large-P continuous simulations. Each statistic plotted is an average over 10 runs and is the point selected by maximizing the F1Adj score.}
    \label{fig:evaluation_large_p_continuous}
\end{figure}

Figures \ref{fig:evaluation_large_n-continuous} present AP, AR, AHP, and AHR as a function of sample size for continuous and mixed data cases. Figure \ref{fig:evaluation_large_p_continuous} shows single run results for large p and small N; in particular, for 50 nodes average degree 4 and 100 nodes average degree 4, for BOSS/BF-BIC and PC-Max/BF-LRT. Full result tables are available on our GitHub site. The timing results are given in Appendix Section~\ref{appendix:simulation_details}.

\section{Simulation Discussion}
\label{sec:discussion}

We now discuss the results of the simulated data analysis.

\paragraph{Small-scale simulations.}  
For continuous data (Figure~\ref{fig:continuous_small_scale}), adjacency precision (AP) is consistently high ($>0.85$), with BOSS/BF-BIC often exceeding $0.95$. Kernel-based PC-Max/KCI performs well for 10-node graphs but is slow and fails to complete in larger cases, while PC-Max/RCIT lags in denser graphs (average degree 4). Adjacency recall (AR) is strong across all methods, with KCI excelling in the smallest cases and BF-based methods maintaining accuracy as sample size grows.  

In edge orientation, BOSS/BF-BIC generally leads in arrowhead precision (AHP), while PC-Max/BF-LRT improves slightly in denser graphs. For arrowhead recall (AHR), BOSS/BF-BIC consistently matches or exceeds PC-Max/BF-LRT. KCI and RCIT lag behind on both measures. Mixed-variable results (Figure~\ref{fig:mixed_small_scale}) follow the same patterns, though AHR is uniformly lower.

\paragraph{Large-$N$ simulations.}  
For very large samples (Figure~\ref{fig:evaluation_large_n-continuous}), PC-Max/BF-LRT tends to introduce too many adjacencies (lower AP) at $N=100{,}000$, though AR remains near-optimal. BOSS/BF-BIC strongly outperforms PC-Max/BF-LRT in AHP, while the two methods converge in AHR at the largest sample size.

\paragraph{Large-$p$ simulations.}  
For higher-dimensional graphs (Figure~\ref{fig:evaluation_large_p_continuous}), PC-Max/BF-LRT attains near-perfect AP and AR, but struggles with orientation (low AHP and AHR) in the 50-node average-degree-2 case. By contrast, BOSS/BF-BIC excels at orientation but performs slightly worse in adjacency detection. This suggests PC-Max/BF-LRT may be preferable when adjacency structure is the main goal, while BOSS/BF-BIC is better suited when accurate orientation is critical.

\paragraph{Tuning considerations.}  
Both methods benefit from larger truncation limits and stronger regularization as $N$ increases. We expanded parameter ranges accordingly, though more targeted tuning could further improve performance. In these experiments, parameter choice was guided by the F1Adj score.

\paragraph{Limitations and robustness.}  
Theoretical guarantees for BF-BIC and BF-LRT rely on additive nonlinear relationships and exponential-family noise. Real-world data may exhibit strong interactions or irregular noise (e.g., heavy-tailed, skewed, multimodal), in which case the approximated likelihood may be misspecified. Kernel-based tests such as KCI and RCIT, though slower, make weaker assumptions and may offer robustness in such scenarios.  

Despite these caveats, the BF-based methods outperformed kernel methods in our simulations, even under fully nonlinear neural causal models (NCMs) with interaction terms. This suggests that truncated additive approximations, combined with regularization, are often sufficient in practice.

\paragraph{Future directions.}  
Future work could explore hybrid approaches that combine basis-expansion efficiency with kernel flexibility, or diagnostics for detecting when additive approximations break down. Our comparisons focused on kernel-based CI tests due to computational limits, but similar trade-offs could be studied against other nonlinear methods such as CAM, GAM-GES, or deep generative approaches.

\section{A Real Data Example}
\label{real_data_example}

We analyze the Algerian Forest Fire dataset from the UCI Irvine Repository \citep{algerian_forest_fires_547}. 

For consistency with the simulation results, we apply the BOSS algorithm with the BF-BIC score, under the assumption that latent variables do not exist. For the sample size $N=244$, our simulation study (Appendix Section \ref{simulation}, Table~1) suggests that a truncation limit of 3 with penalty discount 1 is effective.

To incorporate domain knowledge, we impose background tiering constraints. Specifically, the indices of the Canadian Fire System cannot cause \textit{Region}, \textit{Day}, \textit{Month}, or the weather variables (Relative Humidity, Rain, Temperature, Wind Speed). Similarly, weather variables cannot causally influence \textit{Region}, \textit{Day}, or \textit{Month}. We therefore arrange the variables into four tiers, grouping the deterministic indices following \citet{li2024causal}:\footnote{Following Li et al., who establish correctness of DGES under determinism, we note that substituting a faithfulness-free score-based search such as BOSS for the GES + exact-search combination preserves the same theoretical justification, since both rely on SMR rather than Faithfulness. The algorithm contains three phases; we contend that running BOSS here instead of GES, knowing already which variables are deterministic functions of their parents and clustering them together in the search, allows us to apply the DGES theory directly for all three phases. The reason is that BOSS, like the exact search they are appealing to for the same reason, does not assume Faithfulness, so we do not need to use exact search in Phase 3 to achieve this end. Thus, Phase 1 is satisfied since we are applying a scoring algorithm, and Phase 2 is satisfied since we've already identified and clustered the deterministic variables, by taking account of how they are calculated. Since the reason for including Phase 3 has already been covered by using BOSS instead of GES, we conjecture that, by analogy to DGES, BOSS should yield a correct model in principle.}

\begin{samepage}
\begin{enumerate}
    \item Region, Day, Month
    \item Relative Humidity (RH), Rain, Temperature, Wind Speed (Ws)
    \item BUI (Build-Up Index), DC (Drought Code), DMC (Duff Moisture Code), FFMC (Fine Fuel Moisture Code), FWI (Fire Weather Index), ISI (Initial Spread Index)
    \item Fire
\end{enumerate}
\end{samepage}

We present two results: one restricted to measured variables only, and one including the deterministic indices from the Canadian Fire Weather Index (FWI) system. 

\begin{figure}
    \centering
    \includegraphics[width=0.5\linewidth]{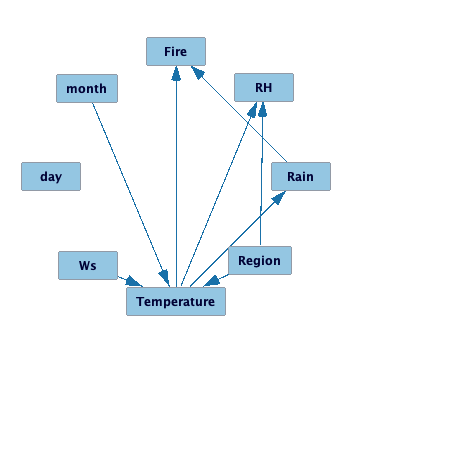}
    \caption{BOSS/BF-BIC CPDAG using measured variables only.}
    \label{fig:boss-bfs-measures-only}
\end{figure}

\paragraph{(1) Measured Variables Only.}  
In the measured-variable model (Figure~\ref{fig:boss-bfs-measures-only}), the weather covariates—RH, Temperature, Rain, and Wind Speed—emerge as direct or partially oriented predictors of \textit{Fire}. These links are consistent with fire science: higher temperatures and lower RH increase fuel dryness and ignition risk, while rainfall suppresses ignition. The direct effect of RH on \textit{Fire} is especially plausible given its strong influence on fuel moisture. Temperature also drives Rain and Ws, consistent with Mediterranean-type meteorological correlations. Finally, \textit{Day} and \textit{Region} act as partial predictors of \textit{Fire}, reflecting seasonal and regional variability. Overall, even without indices, the model recovers interpretable and plausible fire-related mechanisms.

\begin{figure}
    \centering
    \includegraphics[width=0.5\linewidth]{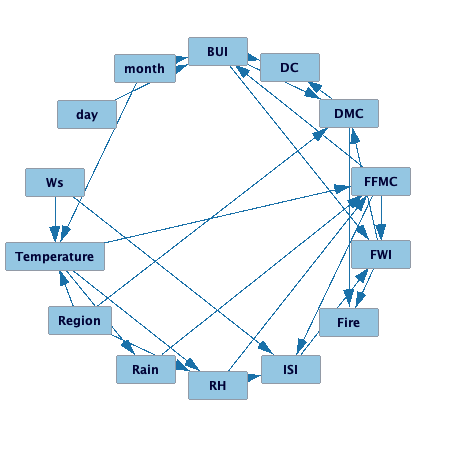}
    \caption{BOSS/BF-BIC CPDAG including Canadian Fire System indices.}
    \label{fig:boss-bfs-all-vars}
\end{figure}

\paragraph{(2) All Variables, Including Deterministic Indices.}  
When the FWI components are included (Figure~\ref{fig:boss-bfs-all-vars}), the model identifies both direct and indirect causal paths from weather variables and indices to \textit{Fire}. The edge \( \text{FWI} \rightarrow \text{Fire} \) is expected, as FWI is designed as a consolidated risk measure. Other plausible relationships include \( \text{DMC} \rightarrow \text{Fire} \) and \( \text{FFMC} \rightarrow \text{Fire} \), since these indices quantify fuel moisture conditions central to ignition. The model also recovers known relationships among indices—for example, \( \text{FFMC} \rightarrow \text{ISI} \rightarrow \text{FWI} \)—reflecting the calculation pipeline of the FWI system. Some mild edge reversals (e.g., \( \text{FWI} \rightarrow \text{DMC} \)) likely result from deterministic overlap in inputs rather than genuine causal reversal. From an applied standpoint, the inclusion of deterministic indices improves interpretability by recovering the structure of the FWI system and clarifying how meteorological variables propagate through indices to affect fire risk.

\noindent
\textbf{Summary.}  
The model with only measured variables captures core weather and spatiotemporal effects on fire, while the full model including deterministic indices yields a richer and more interpretable structure. It recovers known relationships among the indices and confirms the role of FWI and related components as key predictors of fire occurrence. A more detailed expert interpretation, including comparisons to DG, KCI, and RCIT, is provided in Appendix~\ref{appendix:expert_fire_analysis}.

\subsection{Galaxy Analysis}

Beyond this environmental application, FCIT and the associated basis-function methods have recently been used in astrophysical research at much larger scales.  \citet{desmond2025causalstructuregalacticastrophysics} applied FCIT \citet{ramsey2025efficientlatentvariablecausal} with both the Basis-Function LRT and Basis-Function BIC to a sample of roughly $5\times10^5$ low-redshift galaxies from the NASA Sloan Atlas, recovering a hierarchical, mass-driven causal structure among galaxy properties while distinguishing these relations from observational selection effects.  This study, submitted to MNRAS, demonstrates that our nonlinear score and test framework scales effectively to astronomical sample sizes and yields interpretable causal hypotheses in complex physical systems.

\section{Conclusion}
\label{conclusion}

We have demonstrated that BF-based methods combine computational efficiency with accuracy, outperforming kernel-based approaches in scalability while maintaining strong adjacency and orientation recovery for nonlinear data. In high-dimensional settings, BOSS/BF-BIC proved especially effective for identifying causal directionality, while PC-Max/BF-LRT excelled in detecting adjacency structures.

In our real-data analysis of the Algerian Forest Fire dataset, we applied the BOSS \citep{andrews2023fast} algorithm and recovered plausible causal structures in the form of Partial Ancestral Graphs (PAGs). In the full model, which included deterministic Canadian Fire System indices, FCIT correctly identified the expected edge \( \text{FWI} \rightarrow \text{Fire} \) along with much of the computational structure underlying fire risk indices. The measured-only model also recovered interpretable predictors such as RH and Temperature, reinforcing the reliability of our approach. These results suggest that combining FCIT with BF-based scores and tests yields informative models even in the presence of nonlinearities and latent confounding.

Although our theoretical justification for BF-BIC and BF-LRT is grounded in additive and post-nonlinear structural models with exponential-family residuals, simulations show that both methods perform well even when the data-generating process is more general—fully nonlinear, including interaction terms, and not strictly post-nonlinear. This suggests that while the formal theory secures consistency in the additive/post-nonlinear case, the methods extend in practice to a broader class of additive-noise models, providing a useful balance between rigor and applicability. Clarifying the limits of this robustness—both empirically and theoretically—remains an important direction for future work.

Further development of latent-variable search under nonlinear models is also warranted. This includes theoretical analysis of identifiability under nonlinear and non-Gaussian settings, and empirical evaluation across diverse real-world datasets. Refining model diagnostics and relaxing assumptions on functional forms and noise distributions may further improve robustness in less idealized applications.\footnote{The \texttt{py-tetrad} package, including Python scripts and examples, is publicly available at \url{https://github.com/cmu-phil/py-tetrad}. BF-BIC, BF-LRT, DG-BIC, DG-LRT, BOSS, and PC-Max are implemented in Java within the Tetrad project (\url{https://github.com/cmu-phil/tetrad}), with integration into Python via \textit{py-tetrad} and R via \textit{rpy-tetrad}, \url{https://github.com/cmu-phil/py-tetrad/tree/main/pytetrad/R}. The project repository with seeds, true and estimated DAGs, and all scripts for reproducing simulation and real-data experiments is at \url{https://github.com/cmu-phil/bftools}.}

Finally, a natural extension of our framework is to incorporate multiplicative terms involving discrete exogenous variables to capture structural heterogeneity across regimes. For instance, basis functions could be interacted with indicators for region, time, or experimental condition, allowing functional relationships to vary systematically with context. While not pursued here, this extension offers a principled way to model context-specific dependencies and may enhance the flexibility of basis-expansion methods in applications where regime-specific behavior is expected or scientifically meaningful.

\begin{ack}
    Thanks to Clark Glymour for helpful comments on earlier drafts.

    BA was supported by the US National Institutes of Health under the Comorbidity: Substance Use Disorders and Other Psychiatric Conditions Training Program T32DA037183. JR was supported by the US Department of Defense under Contract Number FA8702-15-D-0002 with Carnegie Mellon University for the operation of the Software Engineering Institute. PS was supposed by NIH Award Number: 1043252, ``Interpretable graphical models for large multi-model COPD data'' and NSF award number 2229881, ``AI Institute for Societal Decision Making.'' The content of this paper is solely the responsibility of the authors and does not necessarily represent the official views of these funding agencies. The authors have no conflict of interest to report.
\end{ack}
 
\newpage 

\bibliographystyle{apalike}
\bibliography{refs.bib} 

\begin{thebibliography}{}

\bibitem[Abid, 2019]{algerian_forest_fires_547}
Abid, F. (2019).
\newblock {Algerian Forest Fires}.
\newblock UCI Machine Learning Repository.
\newblock {DOI}: https://doi.org/10.24432/C5KW4N.

\bibitem[Andrews et~al., 2019]{andrews2019learning}
Andrews, B., Ramsey, J., and Cooper, G.~F. (2019).
\newblock Learning high-dimensional directed acyclic graphs with mixed data-types.
\newblock In {\em The 2019 ACM SIGKDD Workshop on Causal Discovery}, pages 4--21. PMLR.

\bibitem[Andrews et~al., 2023]{andrews2023fast}
Andrews, B., Ramsey, J., Sanchez~Romero, R., Camchong, J., and Kummerfeld, E. (2023).
\newblock Fast scalable and accurate discovery of dags using the best order score search and grow shrink trees.
\newblock {\em Advances in Neural Information Processing Systems}, 36:63945--63956.

\bibitem[Brown, 1986]{brown1986fundamentals}
Brown, L.~D. (1986).
\newblock {\em Fundamentals of Statistical Exponential Families: with Applications in Statistical Decision Theory}, volume~9.
\newblock Institute of Mathematical Statistics.

\bibitem[B{\"u}hlmann et~al., 2014]{buhlmann2014cam}
B{\"u}hlmann, P., Peters, J., and Ernest, J. (2014).
\newblock Cam: Causal additive models, high-dimensional order search and penalized regression.
\newblock {\em Annals of Statistics}, 42(6):2526--2556.

\bibitem[Chickering, 2002]{chickering2002optimal}
Chickering, D.~M. (2002).
\newblock Optimal structure identification with greedy search.
\newblock {\em Journal of machine learning research}, 3(Nov):507--554.

\bibitem[Cybenko, 1989]{cybenko1989approximation}
Cybenko, G. (1989).
\newblock Approximation by superpositions of a sigmoidal function.
\newblock {\em Mathematics of control, signals and systems}, 2(4):303--314.

\bibitem[Desmond and Ramsey, 2025]{desmond2025causalstructuregalacticastrophysics}
Desmond, H. and Ramsey, J. (2025).
\newblock The causal structure of galactic astrophysics.

\bibitem[Goudet et~al., 2018]{goudet2018learning}
Goudet, O., Kalainathan, D., Caillou, P., Guyon, I., Lopez-Paz, D., and Sebag, M. (2018).
\newblock Learning functional causal models with generative neural networks.
\newblock {\em Explainable and interpretable models in computer vision and machine learning}, pages 39--80.

\bibitem[Haughton, 1988]{haughton1988choice}
Haughton, D.~M. (1988).
\newblock On the choice of a model to fit data from an exponential family.
\newblock {\em The annals of statistics}, pages 342--355.

\bibitem[He et~al., 2015]{he2015delving}
He, K., Zhang, X., Ren, S., and Sun, J. (2015).
\newblock Delving deep into rectifiers: Surpassing human-level performance on imagenet classification.
\newblock In {\em Proceedings of the IEEE international conference on computer vision}, pages 1026--1034.

\bibitem[Hoyer et~al., 2009]{hoyer2009nonlinear}
Hoyer, P.~O., Janzing, D., Mooij, J.~M., Peters, J., and Sch{\"o}lkopf, B. (2009).
\newblock Nonlinear causal discovery with additive noise models.
\newblock {\em Advances in Neural Information Processing Systems}, 21.

\bibitem[Huang et~al., 2020]{huang2020generalized}
Huang, B., Zhang, K., and Zhang, J. (2020).
\newblock Generalized score-based causal structure learning.
\newblock In {\em Advances in Neural Information Processing Systems (NeurIPS)}, volume~33, pages 5709--5721.

\bibitem[Lam et~al., 2022]{lam2022greedy}
Lam, W.-Y., Andrews, B., and Ramsey, J. (2022).
\newblock Greedy relaxations of the sparsest permutation algorithm.
\newblock In {\em Uncertainty in Artificial Intelligence}, pages 1052--1062. PMLR.

\bibitem[Lee and Hastie, 2015]{lee2015learning}
Lee, J.~D. and Hastie, T.~J. (2015).
\newblock Learning the structure of mixed graphical models.
\newblock {\em Journal of Computational and Graphical Statistics}, 24(1):230--253.

\bibitem[Li et~al., 2024]{li2024causal}
Li, L., Dai, H., Al~Ghothani, H., Huang, B., Zhang, J., Harel, S., Bentwich, I., Chen, G., and Zhang, K. (2024).
\newblock On causal discovery in the presence of deterministic relations.
\newblock {\em Advances in Neural Information Processing Systems}, 37:130920--130952.

\bibitem[Louizos et~al., 2017]{louizos2017causal}
Louizos, C., Shalit, U., Mooij, J.~M., Sontag, D., Zemel, R., and Welling, M. (2017).
\newblock Causal effect inference with deep latent-variable models.
\newblock In {\em Advances in Neural Information Processing Systems (NeurIPS)}, volume~30.

\bibitem[Maas et~al., 2013]{maas2013rectifier}
Maas, A.~L., Hannun, A.~Y., Ng, A.~Y., et~al. (2013).
\newblock Rectifier nonlinearities improve neural network acoustic models.
\newblock In {\em Proc. icml}, volume~30, page~3. Atlanta, GA.

\bibitem[McLachlan and Peel, 2000]{mclachlan2000finite}
McLachlan, G.~J. and Peel, D. (2000).
\newblock {\em Finite mixture models}.
\newblock John Wiley \& Sons.

\bibitem[Newey, 1997]{newey1997convergence}
Newey, W.~K. (1997).
\newblock Convergence rates and asymptotic normality for series estimators.
\newblock {\em Journal of Econometrics}, 79(1):147--168.

\bibitem[Peters et~al., 2014]{peters2014causal}
Peters, J., Mooij, J.~M., Janzing, D., and Sch{\"o}lkopf, B. (2014).
\newblock Causal discovery with continuous additive noise models.
\newblock {\em Journal of machine learning research}, 15:2009--205.

\bibitem[Ramsey, 2016]{ramsey2016improving}
Ramsey, J. (2016).
\newblock Improving accuracy and scalability of the pc algorithm by maximizing p-value.
\newblock {\em arXiv preprint arXiv:1610.00378}.

\bibitem[Ramsey and Andrews, 2025]{ramsey2025efficientlatentvariablecausal}
Ramsey, J. and Andrews, B. (2025).
\newblock Efficient latent variable causal discovery: Combining score search and targeted testing.

\bibitem[Ramsey et~al., 2017]{ramsey2017million}
Ramsey, J., Glymour, M., Sanchez-Romero, R., and Glymour, C. (2017).
\newblock A million variables and more: the fast greedy equivalence search algorithm for learning high-dimensional graphical causal models, with an application to functional magnetic resonance images.
\newblock {\em International journal of data science and analytics}, 3:121--129.

\bibitem[Sonoda and Murata, 2017]{sonoda2017neural}
Sonoda, S. and Murata, N. (2017).
\newblock Neural network with unbounded activation functions is universal approximator.
\newblock {\em Applied and Computational Harmonic Analysis}, 43(2):233--268.

\bibitem[Spirtes et~al., 2001]{spirtes2001causation}
Spirtes, P., Glymour, C., and Scheines, R. (2001).
\newblock {\em Causation, prediction, and search}.
\newblock MIT press.

\bibitem[Strobl et~al., 2019]{strobl2019approximate}
Strobl, E.~V., Zhang, K., and Visweswaran, S. (2019).
\newblock Approximate kernel-based conditional independence tests for fast non-parametric causal discovery.
\newblock {\em Journal of Causal Inference}, 7(1):20180017.

\bibitem[Van~Wagner and Pickett, 1987]{vanwagner1987development}
Van~Wagner, C.~E. and Pickett, T.~L. (1987).
\newblock {\em Development and structure of the Canadian Forest Fire Weather Index System}, volume~35 of {\em Forestry Technical Report}.
\newblock Canadian Forestry Service, Government of Canada.

\bibitem[Zhang and Hyv{\"a}rinen, 2009a]{zhang2009identifiability}
Zhang, K. and Hyv{\"a}rinen, A. (2009a).
\newblock On the identifiability of the post-nonlinear causal model.
\newblock In {\em Proceedings of the Twenty-Fifth Conference on Uncertainty in Artificial Intelligence (UAI)}, pages 647--655.

\bibitem[Zhang and Hyv{\"a}rinen, 2009b]{ZhangPNL-paper}
Zhang, K. and Hyv{\"a}rinen, A. (2009b).
\newblock On the identifiability of the post-nonlinear causal model.
\newblock In {\em Proceedings of the Twenty-Fifth Conference on Uncertainty in Artificial Intelligence (UAI-09)}, pages 647--655. AUAI Press.

\bibitem[Zhang and Hyv{\"a}rinen, 2010]{zhang2010nonlinear}
Zhang, K. and Hyv{\"a}rinen, A. (2010).
\newblock A general non-linear non-gaussian causal model.
\newblock In {\em Advances in Neural Information Processing Systems}, volume~23, pages 885--892.

\bibitem[Zhang and Hyvarinen, 2012]{zhang2012identifiability}
Zhang, K. and Hyvarinen, A. (2012).
\newblock On the identifiability of the post-nonlinear causal model.
\newblock {\em arXiv preprint arXiv:1205.2599}.

\bibitem[Zhang et~al., 2012]{zhang2012kernel}
Zhang, K., Peters, J., Janzing, D., and Sch{\"o}lkopf, B. (2012).
\newblock Kernel-based conditional independence test and application in causal discovery.
\newblock {\em arXiv preprint arXiv:1202.3775}.

\bibitem[Zhang et~al., 2016]{zhang2016causal}
Zhang, K., Sch{\"o}lkopf, B., Spirtes, P., and Glymour, C. (2016).
\newblock Causal discovery with deep generative models.

\bibitem[Zhang et~al., 2020]{zhang2020nonlinear}
Zhang, R., Gong, M., Cai, Z., Liu, W., Wang, R., Zhang, K., and Philip, S.~Y. (2020).
\newblock Nonlinear structure learning with additive noise models.
\newblock {\em Proceedings of the 37th International Conference on Machine Learning (ICML)}, pages 11125--11134.

\bibitem[Zheng et~al., 2024]{zheng2024causal}
Zheng, Y., Huang, B., Chen, W., Ramsey, J., Gong, M., Cai, R., Shimizu, S., Spirtes, P., and Zhang, K. (2024).
\newblock Causal-learn: Causal discovery in python.
\newblock {\em Journal of Machine Learning Research}, 25(60):1--8.

\end{thebibliography}

\appendix

\section{Causal Perceptron Network}
\label{appendix:cpn_section}

We simulate from a neural causal model (NCM), implemented as a recursive SEM where each structural function is a multilayer perceptron (MLP); we refer to this variant as the \emph{Causal Perceptron Network (CPN)}.

Elements of this simulation method have been explored previously, notably by \citet{zhang2012identifiability} and \citet{goudet2018learning}. Following these works, our model is designed specifically for recursive Structural Equation Models (rSEMs). Formally, each observed variable \(X_j\) is modeled as
\[
X_j = f_j(\text{pa}(j), e_j),
\]
where \(\text{pa}(j)\) are the parents of node \(j\), \(e_j\) is an independent noise term, and \(f_j\) is an MLP of fixed architecture. For this paper, we assume noise terms are independently and identically distributed (i.i.d.) according to a user-specified distribution and implement the model in PyTorch.

Data points are generated recursively: we first draw random noise samples and then propagate these values in topological order through the neural networks to compute node values. Repeating this process \(N\) times yields an i.i.d.\ sample of size \(N\).

By the Universal Approximation Theorem \citep{cybenko1989approximation}, a sufficiently large MLP with a single hidden layer can approximate any continuous function \citep{goudet2018learning}. We extend this by permitting multiple hidden layers \citep{zhang2012identifiability}, while maintaining a fixed architecture across all structural functions. Each MLP has an input layer equal in size to the number of parent nodes plus one (the additional input being the noise term) and a single output node.

\begin{figure}
    \centering
    \includegraphics[width=0.5\linewidth]{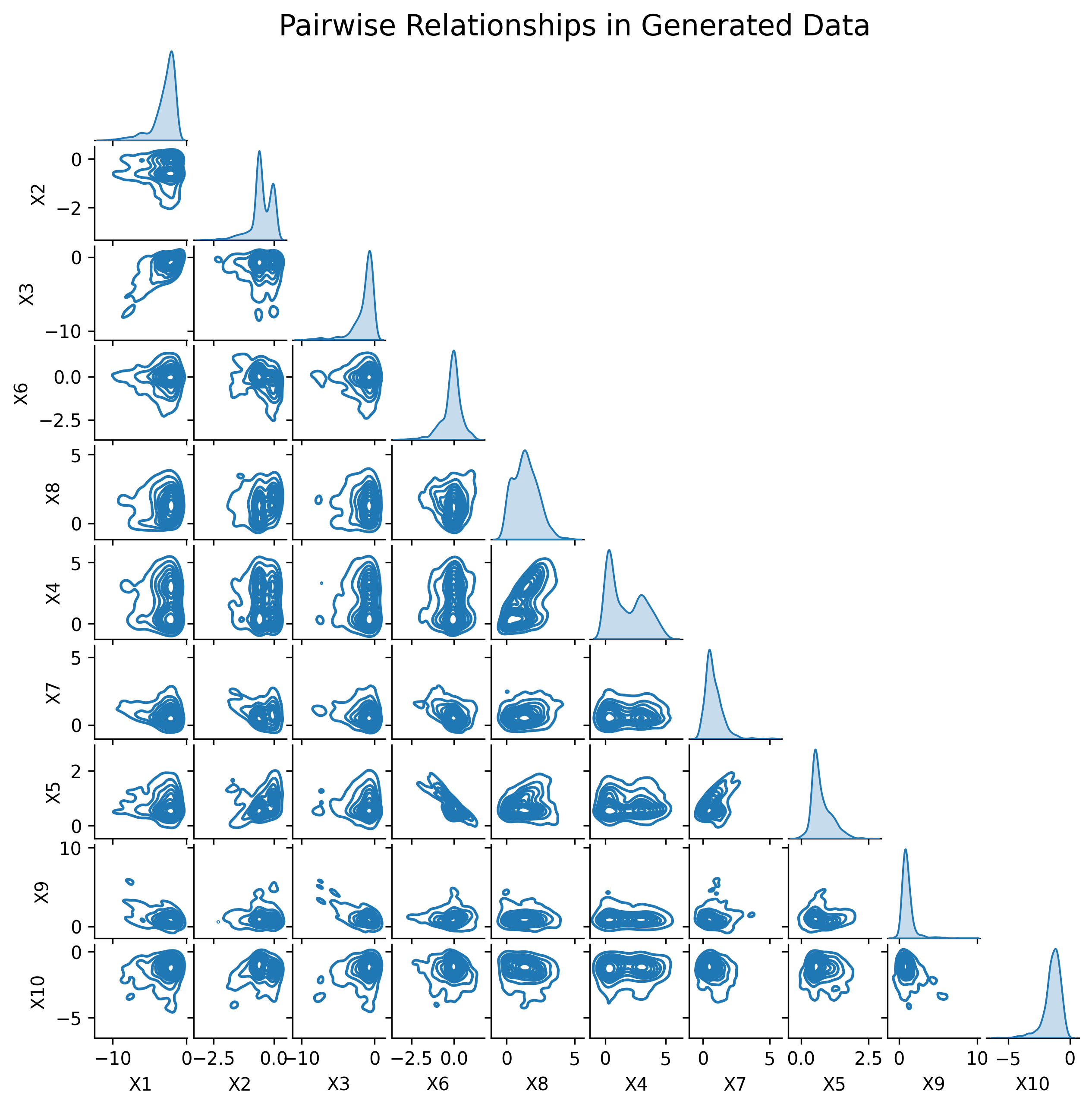}
    \caption{Pairwise plot for an example CPN simulation with 10 nodes, 20 edges, and 1000 samples. Blue contours give a density map using the ``kde'' option in the \texttt{pairplot} method of \texttt{seaborn}.}
    \label{fig:pairwise_plot}
\end{figure}

For initialization, we employ the Leaky Rectified Linear Unit (Leaky ReLU) activation function with input scaling set to 5, Kaiming weight initialization \citep{he2015delving}, and specify \texttt{leaky\_relu} as the nonlinearity. (Universal approximation holds for Leaky ReLU \citep{maas2013rectifier, sonoda2017neural}.) Our default networks contain five hidden layers, each with 50 neurons. The Leaky ReLU function, being piecewise linear, provides flexible responses across the input domain \citep{maas2013rectifier}. These settings enable modeling a broad range of nonlinear functions, producing data distributions similar to those in Figure~\ref{fig:pairwise_plot}.

\paragraph{Multinomial variables.}  
CPN incorporates a parameter, \texttt{multinomial\_prob}, controlling the probability that a variable is multinomial, with between 2 and 5 randomly chosen categories. Setting \texttt{multinomial\_prob} = 0.0 leaves all variables continuous; e.g., \texttt{multinomial\_prob} = 0.2 designates 20\% of variables randomly as multinomial. These variables are inherently generated as categorical rather than discretized post hoc. Multinomial outcomes are sampled using PyTorch’s \texttt{torch.multinomial()} function according to network-inferred probabilities.\footnote{In Python, data are stored in Pandas data frames. The CPN model assigns integer column types for multinomial variables. When data are transferred via JPype to Tetrad, these columns are correctly interpreted as multinomial, ensuring proper handling by BF-BIC and BF-LRT.}

Multinomial nodes are produced by neural networks with a softmax output layer. Given logits \(z_i\), the softmax computes category probabilities
\[
p_i = \frac{e^{z_i}}{\sum_{j} e^{z_j}},
\]
with logits stabilized by subtracting their maximum value. A category is then sampled via \texttt{torch.multinomial()}.

\paragraph{Mixed-data modeling.}  
Multinomial values are treated as continuous inputs for subsequent neural networks, in line with Mixed Graphical Model (MGM) frameworks \citep{lee2015learning} where continuous and discrete variables are modeled jointly. An alternative future extension would be to use one-hot encoding internally, explicitly representing categorical inputs without implicit ordering.

\section{Theoretical Justification}
\label{appendix:theory}

Additive noise models (ANMs) form a class of nonlinear structural equation models in which each variable is modeled as a function of its parents plus independent noise. Early work on ANMs by \citet{hoyer2009nonlinear} and \citet{peters2014causal} demonstrated that under mild assumptions, causal directionality can be inferred from functional asymmetries. \citet{buhlmann2014cam} extended this framework to Causal Additive Models (CAMs), which assume that each structural function is additive over its parents. For theoretical clarity we begin with this additive case, while using basis expansions to approximate the component functions. In Section~\ref{sec:post_nonlin}, we show that the same reasoning extends directly to the more general post-nonlinear (PNL) model, via an invertible transformation.

In contrast to CAM, which uses penalized regression over a space of orders, our BF-BIC approach supports a flexible family of search algorithms and enables the explicit modeling of both continuous and categorical variables. Furthermore, we evaluate the robustness of this additive expansion framework empirically under general (non-additive) structural functions.

BF-BIC extends the Bayesian Information Criterion (BIC) to nonlinear models by embedding each continuous variable into an orthogonal polynomial basis expansion. A central theoretical question is whether this transformation preserves the assumptions required for BIC consistency—particularly the assumption that residuals are distributed according to a known exponential family. This section clarifies the assumptions under which BF-BIC remains theoretically justified and distinguishes between exact and approximate scenarios.

We give a feature comparison of our truncated basis function methods with other additive methods in Table~1, below.

\subsection{Exponential Families and BIC Consistency}

A probability distribution belongs to the exponential family if it has a density (or mass) function of the form:
\[
f(x|\theta) = h(x)\exp\left(\eta(\theta)^\top T(x) - A(\theta)\right),
\]
where \(T(x)\) is a vector of sufficient statistics, \(\eta(\theta)\) the natural parameters, \(h(x)\) the base measure, and \(A(\theta)\) the log-partition function. Gaussian, Poisson, Gamma, Binomial, and Beta distributions are all exponential-family members. 

\citet{haughton1988choice} established that under standard regularity conditions (e.g., smoothness, identifiability, and independent observations), BIC is a consistent model selection criterion when the residuals follow an exponential-family distribution.

\subsection{BF-BIC Consistency with Additive Structural Functions}

\begin{theorem}[BF-BIC Consistency with Additive Structural Functions]
Let \( X = f(\text{pa}(X)) + \varepsilon_X \), where \( \varepsilon_X \) is independent of \( \text{pa}(X) \) and follows an exponential-family distribution. Suppose the structural function \( f \) (or, in the PNL case, the transformed function \( g^{-1}\!\circ f \)) lies in the span of the additive basis functions used (e.g., orthogonal polynomials in each parent variable separately). Then, in the limit of infinite sample size, applying BIC to the expanded model yields consistent selection of the correct parent set.
\end{theorem}

\paragraph{Proof Sketch.} When the basis captures the true function \( f \) exactly, the model is well-specified and falls within the class of exponential-family regressions studied by \citet{haughton1988choice}. Under their regularity conditions, BIC is consistent. \qed

\subsection{Illustrative Example and the Role of Truncation}

Consider the structural equation:
\[
X = Y^3 + YZ + \varepsilon_X, \quad \varepsilon_X \sim \mathcal{N}(0, \sigma^2), \quad \varepsilon_X \ind (Y, Z),
\]
with all variables scaled to lie within \([-1, 1]\). Suppose we expand using Legendre polynomials up to order 3:
\[
Y_1 = Y,\quad Y_2 = \tfrac{1}{2}(3Y^2 - 1),\quad Y_3 = \tfrac{1}{2}(5Y^3 - 3Y).
\]
Then,
\[
X = \tfrac{2}{5}Y_3 + \tfrac{3}{5}Y_1 + (Y_1 Z_1) + \varepsilon_X.
\]

The cross-term \(Y_1 Z_1\) lies outside the additive span. If it is omitted, the residual \( X - X_1 = YZ + \varepsilon_X \) is not Gaussian. Therefore, the residual distribution does not satisfy exponential-family assumptions, and BIC is no longer theoretically guaranteed to be consistent. This underscores the importance of choosing a basis that sufficiently captures the true functional form.

Note that truncated basis expansions yield consistent model selection only if the residuals after expansion are approximately exponential-family. If the residual includes unmodeled structure (e.g., cross-terms or high-order nonlinearities), BIC may fail to select the correct model.

\subsection*{On the Absence of Explicit Interaction Terms}

The basis expansion approach described above restricts attention to additive representations as the working case—i.e., functions of the form \( f(X_1, \dots, X_k) = \sum_i f_i(X_i) \). One might reasonably ask whether the model could be extended to include interaction terms such as \( XY \) or, more generally, products of basis functions across multiple parent variables.

Indeed, it is possible in principle to form such cross-terms by constructing tensor-product bases or directly including interactions like \( \phi_j(X)\psi_k(Y) \). However, such expansions quickly become high-dimensional, particularly in settings with many variables or high polynomial degree, and pose challenges for scalability and model interpretability. More importantly, the resulting conditional distributions often fall outside the exponential-family structure on which our theoretical justification for BIC consistency relies.

Our decision to model only additive terms reflects a trade-off between expressivity and theoretical tractability. Although models like \( X = Y^3 + YZ + \varepsilon \) involve cross-terms, we find that truncated additive expansions often capture such effects heuristically—particularly when the data-generating mechanism includes latent common causes or smooth non-additive dependencies. The additive approximation provides a simple and well-understood statistical structure for analysis, and—via an invertible PNL transform—this same structure can be applied more broadly.

\subsection{Additive Decomposition of BIC}

\begin{theorem}[Additive Decomposition of BF-BIC]
Under the assumptions of Theorem 1 (or the corresponding assumptions after an invertible transform in the PNL setting), if \( X \) is decomposed into orthogonal components \( X_1, X_2, \dots \) via a polynomial expansion, and the basis is sufficient to express \( f \) exactly within \( X_1 \), then:
\[
\text{BIC}(X \mid \text{pa}(X)) = \text{BIC}(X_1 \mid \text{pa}(X_1)),
\]
and additional terms such as \( \text{BIC}(X_2 \mid \cdot) \) do not improve the fit and are penalized by BIC. 
\end{theorem}

\paragraph{Proof Sketch.} BIC is additive across independent components. When \( X_1 \) fully captures the structural function, the remaining variance is pure noise, and further components do not contribute to the likelihood. If the function is only partially captured, higher components help explain additional structure and improve fit. \qed

A natural question is whether one can simply include only the first expanded column of the response variable, \(X_1\), in this way when computing \(\mathrm{BIC}(X \mid Y, Z)\). If the true structural function were indeed perfectly represented by that single term, this reduced model would suffice. In realistic settings, however, the functional dependence of \(X\) on its parents often involves higher-order (or interaction-like) components captured by \(X_2, X_3, \dots\). Omitting these higher-order expansions leaves systematic structure in the residual and undermines the exponential-family assumptions needed for BIC consistency. Empirically, we observe that including the full set of expanded columns \(\{X_1,\dots,X_p\}\) substantially improves adjacency and orientation accuracy, especially for denser graphs or mild deviations from purely additive relationships, while preserving the theoretical convergence guarantees for well-specified expansions.

\subsection{Discrete Variables and the DG Score}

Discrete variables are modeled via the Degenerate Gaussian (DG) Score \citep{andrews2019learning}, which represents each discrete variable as a latent Gaussian variable passed through deterministic thresholds. This latent structure preserves exponential-family assumptions and allows for consistent model selection.

\begin{theorem}[Consistency of DG Score]
Let a discrete variable \(D\) be generated by thresholding a latent Gaussian variable \(D^*\):
\[
D^* \sim \mathcal{N}(\mu, \sigma^2), \quad D = g(D^*).
\]
Then, the likelihood and BIC computed via this representation are asymptotically valid under standard regularity conditions.
\end{theorem}

\paragraph{Proof Sketch.}
This follows from classical results in latent variable modeling and exponential-family mixtures; see \citet{mclachlan2000finite}. \qed

\subsection{Extension to Post-Nonlinear Models}
\label{sec:post_nonlin}

While the preceding discussion assumed that \(X\) follows an additive-nonlinear model 
\[
    X \;=\; f(Y,\,Z,\,e_X),
\]
for additive function $f$, the scope of our approach can be broadened by considering the post-nonlinear (PNL) framework \cite{ZhangPNL-paper}, in which we assume the existence of an arbitrary \emph{invertible} function \(g\) such that
\[
    X \;=\; g\bigl(\tilde{f}(Y,\,Z,\,\tilde{e})\bigr).
\]
Applying \(g^{-1}\) to both sides yields
\[
    g^{-1}(X) \;=\; \tilde{f}(Y,\,Z,\,\tilde{e}),
\]
making \(\tilde{f}\) effectively an additive-nonlinear function. Consequently, 
\[
    X^* := g^{-1}(X)
\]
can be analyzed within the same theoretical framework described for the simpler additive-nonlinear model. 
In particular, one can expand \(X^*\) using the truncated Legendre basis,
\[
    X^* \;=\; \sum_{p=1}^P \beta_p \, \psi_p(Y,\,Z) \;+\; \varepsilon,
\]
where \(\psi_p\) denotes the \(p\)-th Legendre polynomial term (adapted to the dimensions of \(Y\) and \(Z\)) and \(\varepsilon\) is a suitable error term. By treating each coefficient equation (or component) in this expansion separately, the identifiability arguments derived from \cite{haughton1988choice} still apply (provided one can posit exponential or similarly tractable distributions for the noise in each separate component).

Note that in this extended scenario, the overall noise in \(X\)-space need not be purely exponential, because \(X\) itself may be distorting the noise through the invertible function \(g(\cdot)\). Nevertheless, as long as each “component equation” in the \(X^*\)-space maintains an exponential (or other tractable) noise form, the same likelihood-based arguments lead to identifiability. Hence, the BIC-based selection scheme for the basis function parameters retains its validity under the post-nonlinear assumption. See \cite{ZhangPNL-paper} for further theoretical details on the identifiability of post-nonlinear models and \cite{haughton1988choice} for the exponential-based derivations.

This result reveals that the framework originally introduced for the simpler additive-nonlinear case is, in fact, part of a broader class of \emph{post-nonlinear} models. Thus, by inserting the transformation \(g^{-1}\), we can broaden the scope of our analysis while still retaining the same methodological steps of truncated Legendre expansions, exponential-based errors, and BIC-based model selection.

\subsection{Extension to the BF-LRT}

The Basis Function Likelihood Ratio Test (BF-LRT) generalizes the classical likelihood ratio test for conditional independence to the basis-expanded setting. When models are nested and belong to the exponential family, Wilks’ theorem ensures that the likelihood ratio statistic is asymptotically \(\chi^2\)-distributed.

\begin{theorem}[BF-LRT Consistency]
Under nested models in the polynomial-expanded exponential-family framework, the BF-LRT statistic is asymptotically \(\chi^2\)-distributed and provides a consistent test for conditional independence.
\end{theorem}

\paragraph{Proof Sketch.}
This follows from Wilks' theorem for exponential-family models with finite-dimensional parameterizations. The polynomial basis preserves this structure provided sufficient expansion. \qed

\subsection{Conclusion}

BF-BIC and BF-LRT are theoretically consistent when the structural functions lie within the span of the basis functions used. In this case, residuals remain exponential-family, and the standard BIC justification applies. When the expansion is truncated, the method is approximate: BIC and LRT remain useful heuristics but lose strict consistency guarantees. In practice, empirical results indicate that moderate truncation often yields good approximations with strong performance.

\subsection{Comparison of Additive Nonlinear Methods}
\label{appendix:method_comparison}

The following table compares key assumptions and properties of CAM \citep{buhlmann2014cam}, Additive Noise Models (ANMs) \citep{hoyer2009nonlinear}, and our Basis Function approach.

\begin{table}[H]
\centering
\small
\begin{tabular}{p{2.6cm} | p{2.8cm} | p{2.8cm} | p{3.2cm}}
\toprule
\textbf{Feature} & \textbf{CAM (2014)} & \textbf{Additive Noise Models (ANMs)} & \textbf{Our BF-BIC / BF-LRT Approach} \\
\midrule
\textbf{Structure of model} & Additive: \( X_j = \sum f_{jk}(X_k) + \varepsilon_j \) & Additive: \( X_j = f_j(\text{pa}(j)) + \varepsilon_j \) & Additive expansions (orthogonal basis); generalizes to post-nonlinear models via invertible transform \\
\midrule
\textbf{Noise distribution} & Gaussian & Typically non-Gaussian (for identifiability) & Exponential family (e.g., Gaussian, Poisson); preserved under PNL transform \\
\midrule
\textbf{Identifiability guarantee} & From variable ordering under additivity and smoothness & From asymmetry due to non-Gaussianity & From exponential-family structure and sufficient basis expansion; extends to PNL via invertible reparametrization \\
\midrule
\textbf{Function estimation} & Penalized regression (e.g., splines) & Nonparametric (e.g., GP, kernel methods) & Orthogonal polynomial basis (e.g., Legendre) \\
\midrule
\textbf{Handles discrete variables} & Not directly & Not typically & Yes—via DG embedding or hybrid structure \\
\midrule
\textbf{Consistency guarantees} & Yes, under smoothness and sparsity & Yes, under suitable assumptions & Yes, under additive structure or PNL transform with sufficient basis \\
\midrule
\textbf{Supports score-based search} & Yes (order search + regression) & Not typically & Yes (e.g., with BOSS or GES) \\
\midrule
\textbf{Supports constraint-based search} & No & Yes (e.g., with KCI) & Yes (BF-LRT is a CI test) \\
\midrule
\textbf{Captures interactions} & No (additive only) & No (unless modeled explicitly) & Not explicitly — approximated via truncated basis; PNL transform can expose additional structure \\
\midrule
\textbf{Scalability} & Moderate (search over orderings with regression fits) & Often poor (nonparametric CI testing) & High (fixed-size basis + efficient algorithms) \\
\bottomrule
\end{tabular}
\caption{Comparison of additive nonlinear causal discovery methods. The BF-BIC/BF-LRT framework reduces to the additive case but extends directly to the post-nonlinear setting via invertible transformations.}
\label{tab:additive_comparison}
\end{table}

In summary, while our formal proofs establish consistency under additive structural functions, the same reasoning extends naturally to the broader class of post-nonlinear models via reparameterization with an invertible transform. This shows that BF-BIC and BF-LRT are not limited to the additive case but inherit their consistency guarantees under the more general post-nonlinear assumption. Empirically, even with truncated expansions, these methods remain robust and scalable, providing practical accuracy beyond what the strict additive theory would suggest.

\section{Expert Analysis of the Algerian Forest Fire Models}
\label{appendix:expert_fire_analysis}

In this appendix, we provide an expert analysis of the models learned by LV-Lite on the Algerian Forest Fire dataset, using domain knowledge from the Canadian Fire Weather Index (FWI) system. Two models are analyzed: one using only measured variables and one including the deterministic indices from the FWI system. For reference, the formulas defining the Canadian indices are provided earlier in this appendix. The “expert” here is ChatGPT o1 on 2025-04-15, where we requested that the LLM assume the role of an expert in forest-fire data analysis with detailed knowledge of the Canadian Fire System and offer views on which edges might be helpful to a user interested in causal analysis and which may be misleading. We offer this as an experiment in using this LLM to play the role of such an expert, as analyses for real datasets from human experts may be difficult to obtain.

\subsection{Expert Analysis of Alternative Graphs with Emphasis on Potential Misinterpretations}
\label{appendix:expert_fire_analysis_details}

In this subsection, we compare and interpret six alternative graphs learned from the Algerian Forest Fire dataset, each using a different combination of model assumptions (linear vs.\ nonlinear) and different conditional independence tests or scoring criteria (DG-BIC, BF-BIC, KCI, RCIT). We include two main categories of variable sets: (1)~\textbf{measured variables only} (i.e., meteorological factors, \textit{Region}, \textit{Day}, \textit{Month}, and the binary \textit{Fire} indicator) and (2)~\textbf{all variables}, which additionally include the Canadian Fire Weather Index (FWI) deterministic indices (\textit{FFMC}, \textit{DMC}, \textit{DC}, \textit{ISI}, \textit{BUI}, \textit{FWI}). Each learned model is assessed from the perspective of an expert in fire-data analysis, with particular attention to how certain edges might be misleading—especially those pointing into \textit{Fire}.

\begin{figure}
    \centering
    \includegraphics[width=0.5\linewidth]{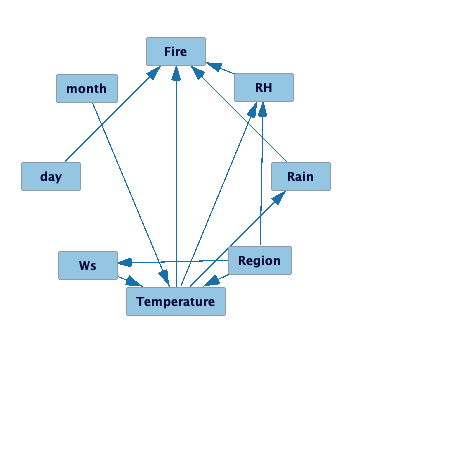}
    \caption{BOSS/DG-BIC CPDAG for measured variables only.}
    \label{fig:boss-dgs-measured}
\end{figure}

\begin{figure}
    \centering
    \includegraphics[width=0.5\linewidth]{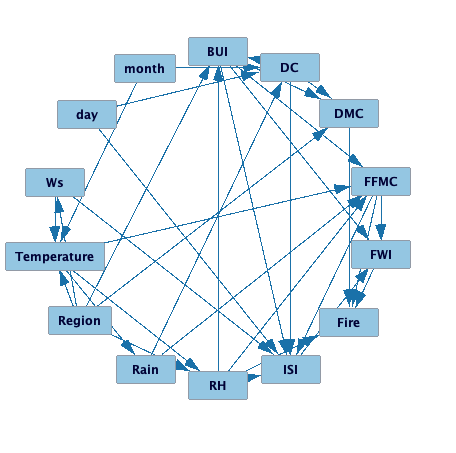}
    \caption{BOSS/DG-BIC CPDAG for all variables including Canadian Fire System indices.}
    \label{fig:boss-dgs-all-vars}
\end{figure}

\begin{figure}
    \centering
    \includegraphics[width=0.5\linewidth]{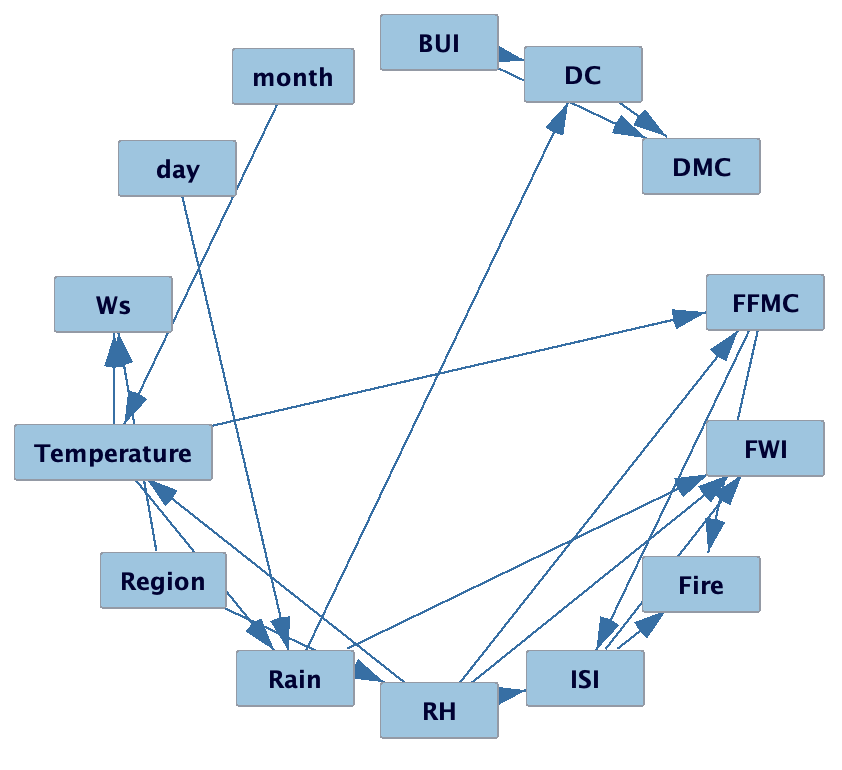}
    \caption{PC/KCI for all variables including Canadian Fire System indices.}
    \label{fig:pc-kci-all-vars}
\end{figure}

\begin{figure}
    \centering
    \includegraphics[width=0.5\linewidth]{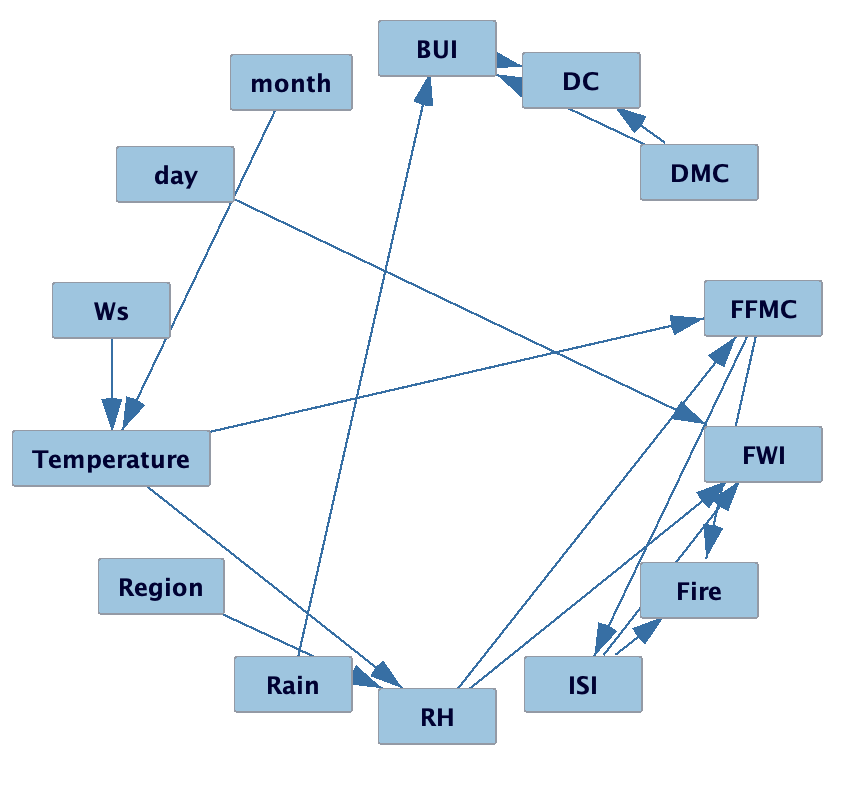}
    \caption{PC/RCIT for all variables including Canadian Fire System indices.}
    \label{fig:pc-rcit-all-vars}
\end{figure}

Two of these models were given in the main text: (1) the BOSS-BF-BIC model on measured variables only (Figure~\ref{fig:boss-bfs-measures-only}) and (3) the BOSS-BF-BIC model on all variables including the Canadian Fire System indices (Figure~\ref{fig:boss-bfs-all-vars}). The remaining are: (2) the BOSS-DG-BIC model on measured variables only (Figure~\ref{fig:boss-dgs-measured}); (4) the BOSS-DG-BIC model on all variables including the Canadian Fire System indices (Figure~\ref{fig:boss-dgs-all-vars}); (5) the PC-KCI model on all variables including the Canadian Fire System indices (Figure~\ref{fig:pc-kci-all-vars}); and (6) the PC-RCIT model on all variables including the Canadian Fire System indices (Figure~\ref{fig:pc-rcit-all-vars}).

\paragraph{1. Measured Linear (BOSS/DG-BIC)}
The first model (Figure~\ref{fig:boss-dgs-measured}) uses the BOSS algorithm under a degenerate Gaussian (DG) scoring criterion on the measured variables only.
\begin{itemize}
    \item \textbf{Interpretable Edges:} This graph highlights \(\text{Temperature} \rightarrow \text{Fire}\), \(\text{RH} \rightarrow \text{Fire}\), and \(\text{Rain} \rightarrow \text{Fire}\), aligning well with domain knowledge that fire occurrence is strongly tied to heat, dryness, and rainfall suppression.
    \item \textbf{Potentially Misleading Aspects:} The model shows \(\text{Temperature} \rightarrow \text{RH}\), which can suggest a causal interpretation that higher temperature \emph{causes} lower relative humidity. In reality, both variables may respond to larger-scale meteorological processes (e.g., seasonal or synoptic patterns), so the edge direction can be overstated. Similarly, the direct link \(\textit{Day} \rightarrow \textit{Fire}\) may over-represent a strictly causal effect, whereas in practice \textit{Day} is more of a proxy for seasonal or monthly variation. Readers should note that in the absence of the Canadian Fire System indices, the model can over-simplify the true pathways by placing undue emphasis on any single meteorological driver.
\end{itemize}

\paragraph{2. Measured Nonlinear (BOSS/BF-BIC)}
The second model (Figure~\ref{fig:boss-bfs-measures-only}) applies BF-BIC (allowing nonlinearities) to the same measured-only variables.
\begin{itemize}
    \item \textbf{Interpretable Edges:} Core weather-to-\textit{Fire} relationships remain (e.g., \(\text{Temperature} \rightarrow \text{Fire}\), \(\text{Rain} \rightarrow \text{Fire}\), \(\text{RH} \rightarrow \text{Fire}\) in some partial forms), consistent with the fact that higher temperatures and lower humidity often increase the risk of ignition, while rainfall can mitigate it.
    \item \textbf{Potentially Misleading Aspects:} Due to nonlinear scoring, some apparently direct edges might mask confounding or shared-cause relationships. For instance, \(\text{Region} \rightarrow \text{Temperature}\) is plausible in a broad sense (certain regions being warmer), but it can also be a stand-in for larger-scale climatic gradients. Likewise, edge orientation can give the impression of direct causal influence on \textit{Fire}, whereas in reality multiple unmeasured factors (e.g., fuel load, human ignition sources) may also play critical roles.
\end{itemize}

\paragraph{3. All Variables, Linear (BOSS/DG-BIC)}
In the graph of Figure~\ref{fig:boss-dgs-all-vars}, we add the deterministic Canadian Fire System indices under a linear scoring regime.
\begin{itemize}
    \item \textbf{Interpretable Edges:} We see many expected structures, such as \(\text{BUI} \rightarrow \text{FWI} \rightarrow \text{Fire}\) and \(\text{FFMC} \rightarrow \text{Fire}\). This reflects the domain truth that the sub-indices are calculated from weather inputs and strongly predict fire occurrence.
    \item \textbf{Potentially Misleading Aspects:} Because the FWI components are partly deterministic transformations of meteorological inputs, edges like \(\text{RH} \rightarrow \text{Fire}\) may be overstated. In many real operational contexts, \(\text{RH}\) influences \textit{Fire} \emph{through} the sub-indices (e.g., \(\text{FFMC}\)), so the direct arrow to \textit{Fire} might represent correlation rather than a genuine unmediated cause. Moreover, certain reversed edges (e.g., \(\text{FFMC} \rightarrow \text{BUI}\) if it appeared) can be symptomatic of near-deterministic relationships that the linear model cannot fully distinguish.
\end{itemize}

\paragraph{4. All Variables, Nonlinear (BOSS/BF-BIC)}
This model (Figure~\ref{fig:boss-bfs-all-vars}) uses BF-BIC (nonlinear) with the full variable set.
\begin{itemize}
    \item \textbf{Interpretable Edges:} We often observe \(\text{FFMC} \rightarrow \text{FWI} \rightarrow \text{Fire}\), which is precisely the logic of the Canadian system: dryness of fine fuels influences FWI, which is in turn strongly predictive of fire. Additional sub-indices like \(\text{DMC}\) and \(\text{DC}\) can also point toward \textit{Fire}, reflecting deeper fuel-moisture conditions.
    \item \textbf{Potentially Misleading Aspects:} Because of the strong correlations and partial determinism among the indices, edge orientation (e.g., \(\text{FWI} \rightarrow \text{DMC}\)) can reverse or appear illogical from a purely physical standpoint. This doesn’t necessarily mean the model is incorrect; rather, it indicates that the algorithm sees insufficient residual variance to disambiguate directionality among heavily interdependent indices. Analysts should treat such “reversed” edges with caution and rely on domain knowledge of how indices are computed.
\end{itemize}

\paragraph{5. PC/KCI on All Variables}
The output in Figure~\ref{fig:pc-kci-all-vars} employs the PC algorithm with Kernel Conditional Independence (KCI) tests.
\begin{itemize}
    \item \textbf{Interpretable Edges:} PC/KCI frequently retains the high-level chain from weather variables (e.g., \(\text{Temperature}, \text{RH}, \text{Rain}\)) to Canadian Fire System indices, culminating in \(\text{FFMC}, \text{ISI}, \text{FWI} \rightarrow \text{Fire}\). The presence of these edges is consistent with operational knowledge.
    \item \textbf{Potentially Misleading Aspects:} Constraint-based algorithms like PC can omit or incompletely orient edges when near-deterministic relationships are present. Thus, some sub-indices that are strictly computed from others may receive partly unoriented edges or directions that conflict with standard FWI formulas. For instance, \(\text{RH} \rightarrow \text{FWI}\) might appear direct, even though \(\text{FWI}\) is explicitly calculated from intermediate indices. Additionally, the PC approach may split or re-orient edges around \(\textit{Fire}\) if the partial correlations are ambiguous.
\end{itemize}

\paragraph{6. PC/RCIT on All Variables}
Finally, Figure~\ref{fig:pc-rcit-all-vars} applies the PC algorithm with Randomized Conditional Independence Tests (RCIT).
\begin{itemize}
    \item \textbf{Interpretable Edges:} Like the KCI version, we still see key connections from the sub-indices to \textit{Fire}, generally consistent with the design of the Canadian system. Edges such as \(\text{FFMC} \rightarrow \text{Fire}\) or \(\text{ISI} \rightarrow \text{Fire}\) reflect known ignition dynamics.
    \item \textbf{Potentially Misleading Aspects:} RCIT can alter adjacency or orientation decisions relative to KCI, sometimes producing edges such as \(\textit{Day} \rightarrow \text{FWI}\). In practice, we know \textit{Day} itself does not factor into FWI calculations directly—\textit{Day} is more of a proxy for seasonal progression. Treating this as a strict causal link could thus misrepresent the formulaic relationship. Moreover, any edges from weather directly to \textit{Fire} might conflate the fact that the system’s official “cause” is the chain weather \(\rightarrow\) sub-indices \(\rightarrow\) \textit{Fire}.
\end{itemize}

\paragraph{Summary and Recommendations}
From a fire-analysis perspective, each model recovers key drivers of ignition risk, but all can be misleading if interpreted naively:
\begin{itemize}
    \item \textit{Measured-only} models often overstate direct weather-to-fire links (omitting mediating indices) and can attribute causal roles to \textit{Region} or \textit{Day} that are partly confounding proxies.
    \item \textit{All-variable} models better reflect the physical structure of the Canadian system, yet may still show reversed edges among deterministic sub-indices. These flips typically arise from the algorithms’ inability to discriminate direction in near-deterministic chains.
    \item \textit{Nonlinear} criteria (BF-BIC, KCI, RCIT) can better capture complex meteorological relationships, but also risk adding spurious direct edges to \textit{Fire} if they fail to fully account for intermediate indices.
    \item \textit{Constraint-based} vs.\ \textit{score-based} differences can lead to unoriented or contradictory edges, especially under partial determinism. Analysts should not assume that a learned direction always reflects physical causality in the presence of correlated inputs.
\end{itemize}

When using these methods in practice, one must remain vigilant: results that place “cause” directly on any single index or meteorological variable for \textit{Fire} might be oversimplifying. Domain knowledge about how the FWI system is formally computed, as well as local fire ecology and seasonal patterns, is crucial for interpreting—and correctly revising—misleading edges in the learned graphs.

\noindent For definitions and formulas of the Canadian Fire Weather Index components, see, e.g., \citet{vanwagner1987development} and the domain discussion in Appendix~\ref{appendix:expert_fire_analysis}.

\subsection{Simulation Details}
\label{appendix:simulation_details}

\begin{table}[htbp]\centering\scriptsize

\caption{Optimal parameter settings by Sample Size (Small-Scale, Continuous Data)}
\label{tab:small_continuous_optimal_params}
\begin{tabular}{lllrrrr}
\toprule
Scale & Type & Label & sample\_size & trunc\_limit & penalty & alpha \\
\midrule
small & Continuous & BOSS-BF-BIC 10:2 & 200 & 3 & 1.000000 & * \\
small & Continuous & BOSS-BF-BIC 10:2 & 500 & 3 & 1.000000 & * \\
small & Continuous & BOSS-BF-BIC 10:2 & 1000 & 3 & 1.000000 & * \\
small & Continuous & BOSS-BF-BIC 10:2 & 2000 & 3 & 2.000000 & * \\
small & Continuous & BOSS-BF-BIC 10:2 & 5000 & 4 & 2.000000 & * \\
small & Continuous & BOSS-BF-BIC 10:2 & 10000 & 4 & 4.000000 & * \\
small & Continuous & BOSS-BF-BIC 10:4 & 200 & 3 & 1.000000 & * \\
small & Continuous & BOSS-BF-BIC 10:4 & 500 & 3 & 1.000000 & * \\
small & Continuous & BOSS-BF-BIC 10:4 & 1000 & 3 & 1.000000 & * \\
small & Continuous & BOSS-BF-BIC 10:4 & 2000 & 4 & 1.000000 & * \\
small & Continuous & BOSS-BF-BIC 10:4 & 5000 & 3 & 4.000000 & * \\
small & Continuous & BOSS-BF-BIC 10:4 & 10000 & 4 & 4.000000 & * \\
small & Continuous & PC-MAX-BF-LRT 10:2 & 200 & 3 & * & 0.050000 \\
small & Continuous & PC-MAX-BF-LRT 10:2 & 500 & 3 & * & 0.050000 \\
small & Continuous & PC-MAX-BF-LRT 10:2 & 1000 & 3 & * & 0.010000 \\
small & Continuous & PC-MAX-BF-LRT 10:2 & 2000 & 8 & * & 0.001000 \\
small & Continuous & PC-MAX-BF-LRT 10:2 & 5000 & 8 & * & 0.010000 \\
small & Continuous & PC-MAX-BF-LRT 10:2 & 10000 & 8 & * & 0.001000 \\
small & Continuous & PC-MAX-BF-LRT 10:4 & 200 & 3 & * & 0.050000 \\
small & Continuous & PC-MAX-BF-LRT 10:4 & 500 & 4 & * & 0.050000 \\
small & Continuous & PC-MAX-BF-LRT 10:4 & 1000 & 8 & * & 0.050000 \\
small & Continuous & PC-MAX-BF-LRT 10:4 & 2000 & 8 & * & 0.050000 \\
small & Continuous & PC-MAX-BF-LRT 10:4 & 5000 & 8 & * & 0.001000 \\
small & Continuous & PC-MAX-BF-LRT 10:4 & 10000 & 8 & * & 0.001000 \\
small & Continuous & PC-KCI 10:2 & 200 & * & * & 0.050000 \\
small & Continuous & PC-KCI 10:2 & 500 & * & * & 0.050000 \\
small & Continuous & PC-KCI 10:2 & 1000 & * & * & 0.050000 \\
small & Continuous & PC-KCI 10:4 & 200 & * & * & 0.050000 \\
small & Continuous & PC-KCI 10:4 & 500 & * & * & 0.050000 \\
small & Continuous & PC-RCIT 10:2 & 200 & * & * & 0.050000 \\
small & Continuous & PC-RCIT 10:2 & 500 & * & * & 0.050000 \\
small & Continuous & PC-RCIT 10:2 & 1000 & * & * & 0.050000 \\
small & Continuous & PC-RCIT 10:2 & 2000 & * & * & 0.050000 \\
small & Continuous & PC-RCIT 10:2 & 5000 & * & * & 0.050000 \\
small & Continuous & PC-RCIT 10:2 & 10000 & * & * & 0.010000 \\
small & Continuous & PC-RCIT 10:4 & 200 & * & * & 0.050000 \\
small & Continuous & PC-RCIT 10:4 & 500 & * & * & 0.050000 \\
small & Continuous & PC-RCIT 10:4 & 1000 & * & * & 0.050000 \\
small & Continuous & PC-RCIT 10:4 & 2000 & * & * & 0.050000 \\
small & Continuous & PC-RCIT 10:4 & 5000 & * & * & 0.050000 \\
small & Continuous & PC-RCIT 10:4 & 10000 & * & * & 0.050000 \\
small & Continuous & BOSS-BF-BIC 20:2 & 200 & 3 & 1.000000 & * \\
small & Continuous & BOSS-BF-BIC 20:2 & 500 & 3 & 1.000000 & * \\
small & Continuous & BOSS-BF-BIC 20:2 & 1000 & 3 & 1.000000 & * \\
small & Continuous & BOSS-BF-BIC 20:2 & 2000 & 3 & 2.000000 & * \\
small & Continuous & BOSS-BF-BIC 20:2 & 5000 & 3 & 4.000000 & * \\
small & Continuous & BOSS-BF-BIC 20:2 & 10000 & 4 & 8.000000 & * \\
small & Continuous & BOSS-BF-BIC 20:4 & 200 & 3 & 1.000000 & * \\
small & Continuous & BOSS-BF-BIC 20:4 & 500 & 3 & 1.000000 & * \\
small & Continuous & BOSS-BF-BIC 20:4 & 1000 & 3 & 1.000000 & * \\
small & Continuous & BOSS-BF-BIC 20:4 & 2000 & 3 & 1.000000 & * \\
small & Continuous & BOSS-BF-BIC 20:4 & 5000 & 3 & 4.000000 & * \\
small & Continuous & BOSS-BF-BIC 20:4 & 10000 & 3 & 8.000000 & * \\
small & Continuous & PC-MAX-BF-LRT 20:2 & 200 & 3 & * & 0.050000 \\
small & Continuous & PC-MAX-BF-LRT 20:2 & 500 & 8 & * & 0.050000 \\
small & Continuous & PC-MAX-BF-LRT 20:2 & 1000 & 8 & * & 0.050000 \\
small & Continuous & PC-MAX-BF-LRT 20:2 & 2000 & 8 & * & 0.010000 \\
small & Continuous & PC-MAX-BF-LRT 20:2 & 5000 & 8 & * & 0.001000 \\
small & Continuous & PC-MAX-BF-LRT 20:2 & 10000 & 8 & * & 0.001000 \\
small & Continuous & PC-MAX-BF-LRT 20:4 & 200 & 3 & * & 0.050000 \\
small & Continuous & PC-MAX-BF-LRT 20:4 & 500 & 3 & * & 0.050000 \\
small & Continuous & PC-MAX-BF-LRT 20:4 & 1000 & 4 & * & 0.050000 \\
small & Continuous & PC-MAX-BF-LRT 20:4 & 2000 & 8 & * & 0.050000 \\
small & Continuous & PC-MAX-BF-LRT 20:4 & 5000 & 8 & * & 0.001000 \\
small & Continuous & PC-MAX-BF-LRT 20:4 & 10000 & 8 & * & 0.001000 \\
\bottomrule
\end{tabular}

\end{table}

\begin{table}[htbp]\centering\scriptsize

\caption{Optimal parameter settings by Sample Size (Small-Scale, Mixed Data)}
\label{tab:small_mixed_optimal_params}
\begin{tabular}{lllrrrr}
\toprule
Scale & Type & Label & sample\_size & trunc\_limit & penalty & alpha \\
\midrule
small & Mixed & BOSS-BF-BIC 10:2 & 200 & 3 & 1.000000 & * \\
small & Mixed & BOSS-BF-BIC 10:2 & 500 & 3 & 1.000000 & * \\
small & Mixed & BOSS-BF-BIC 10:2 & 1000 & 3 & 1.000000 & * \\
small & Mixed & BOSS-BF-BIC 10:2 & 2000 & 3 & 2.000000 & * \\
small & Mixed & BOSS-BF-BIC 10:2 & 5000 & 3 & 2.000000 & * \\
small & Mixed & BOSS-BF-BIC 10:2 & 10000 & 3 & 4.000000 & * \\
small & Mixed & BOSS-BF-BIC 10:4 & 200 & 3 & 1.000000 & * \\
small & Mixed & BOSS-BF-BIC 10:4 & 500 & 3 & 1.000000 & * \\
small & Mixed & BOSS-BF-BIC 10:4 & 1000 & 3 & 1.000000 & * \\
small & Mixed & BOSS-BF-BIC 10:4 & 2000 & 4 & 1.000000 & * \\
small & Mixed & BOSS-BF-BIC 10:4 & 5000 & 3 & 2.000000 & * \\
small & Mixed & BOSS-BF-BIC 10:4 & 10000 & 8 & 1.000000 & * \\
small & Mixed & PC-MAX-BF-LRT 10:2 & 200 & 4 & * & 0.050000 \\
small & Mixed & PC-MAX-BF-LRT 10:2 & 500 & 1 & * & 0.050000 \\
small & Mixed & PC-MAX-BF-LRT 10:2 & 1000 & 3 & * & 0.010000 \\
small & Mixed & PC-MAX-BF-LRT 10:2 & 2000 & 4 & * & 0.050000 \\
small & Mixed & PC-MAX-BF-LRT 10:2 & 5000 & 3 & * & 0.001000 \\
small & Mixed & PC-MAX-BF-LRT 10:2 & 10000 & 8 & * & 0.001000 \\
small & Mixed & PC-MAX-BF-LRT 10:4 & 200 & 4 & * & 0.050000 \\
small & Mixed & PC-MAX-BF-LRT 10:4 & 500 & 4 & * & 0.050000 \\
small & Mixed & PC-MAX-BF-LRT 10:4 & 1000 & 3 & * & 0.050000 \\
small & Mixed & PC-MAX-BF-LRT 10:4 & 2000 & 4 & * & 0.050000 \\
small & Mixed & PC-MAX-BF-LRT 10:4 & 5000 & 8 & * & 0.010000 \\
small & Mixed & PC-MAX-BF-LRT 10:4 & 10000 & 8 & * & 0.010000 \\
small & Mixed & PC-KCI 10:2 & 200 & * & * & 0.050000 \\
small & Mixed & PC-KCI 10:2 & 500 & * & * & 0.001000 \\
small & Mixed & PC-KCI 10:2 & 1000 & * & * & 0.050000 \\
small & Mixed & PC-KCI 10:4 & 200 & * & * & 0.010000 \\
small & Mixed & PC-KCI 10:4 & 500 & * & * & 0.010000 \\
small & Mixed & PC-RCIT 10:2 & 200 & * & * & 0.050000 \\
small & Mixed & PC-RCIT 10:2 & 500 & * & * & 0.001000 \\
small & Mixed & PC-RCIT 10:2 & 1000 & * & * & 0.050000 \\
small & Mixed & PC-RCIT 10:2 & 2000 & * & * & 0.050000 \\
small & Mixed & PC-RCIT 10:2 & 5000 & * & * & 0.050000 \\
small & Mixed & PC-RCIT 10:2 & 10000 & * & * & 0.050000 \\
small & Mixed & PC-RCIT 10:4 & 200 & * & * & 0.050000 \\
small & Mixed & PC-RCIT 10:4 & 500 & * & * & 0.050000 \\
small & Mixed & PC-RCIT 10:4 & 1000 & * & * & 0.050000 \\
small & Mixed & PC-RCIT 10:4 & 2000 & * & * & 0.050000 \\
small & Mixed & PC-RCIT 10:4 & 5000 & * & * & 0.050000 \\
small & Mixed & PC-RCIT 10:4 & 10000 & * & * & 0.050000 \\
small & Mixed & BOSS-BF-BIC 20:2 & 200 & 3 & 1.000000 & * \\
small & Mixed & BOSS-BF-BIC 20:2 & 500 & 3 & 1.000000 & * \\
small & Mixed & BOSS-BF-BIC 20:2 & 1000 & 3 & 2.000000 & * \\
small & Mixed & BOSS-BF-BIC 20:2 & 2000 & 3 & 1.000000 & * \\
small & Mixed & BOSS-BF-BIC 20:2 & 5000 & 3 & 1.000000 & * \\
small & Mixed & BOSS-BF-BIC 20:2 & 10000 & 3 & 2.000000 & * \\
small & Mixed & BOSS-BF-BIC 20:4 & 200 & 3 & 1.000000 & * \\
small & Mixed & BOSS-BF-BIC 20:4 & 500 & 3 & 1.000000 & * \\
small & Mixed & BOSS-BF-BIC 20:4 & 1000 & 3 & 1.000000 & * \\
small & Mixed & BOSS-BF-BIC 20:4 & 2000 & 3 & 1.000000 & * \\
small & Mixed & BOSS-BF-BIC 20:4 & 5000 & 4 & 1.000000 & * \\
small & Mixed & BOSS-BF-BIC 20:4 & 10000 & 4 & 2.000000 & * \\
small & Mixed & PC-MAX-BF-LRT 20:2 & 200 & 3 & * & 0.050000 \\
small & Mixed & PC-MAX-BF-LRT 20:2 & 500 & 3 & * & 0.050000 \\
small & Mixed & PC-MAX-BF-LRT 20:2 & 1000 & 8 & * & 0.050000 \\
small & Mixed & PC-MAX-BF-LRT 20:2 & 2000 & 8 & * & 0.050000 \\
small & Mixed & PC-MAX-BF-LRT 20:2 & 5000 & 8 & * & 0.010000 \\
small & Mixed & PC-MAX-BF-LRT 20:2 & 10000 & 8 & * & 0.001000 \\
small & Mixed & PC-MAX-BF-LRT 20:4 & 200 & 4 & * & 0.050000 \\
small & Mixed & PC-MAX-BF-LRT 20:4 & 500 & 4 & * & 0.050000 \\
small & Mixed & PC-MAX-BF-LRT 20:4 & 1000 & 4 & * & 0.050000 \\
small & Mixed & PC-MAX-BF-LRT 20:4 & 2000 & 8 & * & 0.050000 \\
small & Mixed & PC-MAX-BF-LRT 20:4 & 5000 & 8 & * & 0.001000 \\
small & Mixed & PC-MAX-BF-LRT 20:4 & 10000 & 8 & * & 0.001000 \\
\bottomrule
\end{tabular}

\end{table}

\begin{table}[htbp]\centering\scriptsize

\caption{Optimal parameter settings by Sample Size (Large-N, Continuous Data)}
\label{tab:large_n_continuous_optimal_params}
\begin{tabular}{lllrrrr}
\toprule
Scale & Type & Label & sample\_size & trunc\_limit & penalty & alpha \\
\midrule
large & Continuous & BOSS-BF-BIC Continuous 10:2 & 10000 & 4 & 8.000000 & * \\
large & Continuous & BOSS-BF-BIC Continuous 10:2 & 20000 & 6 & 4.000000 & * \\
large & Continuous & BOSS-BF-BIC Continuous 10:2 & 50000 & 7 & 8.000000 & * \\
large & Continuous & BOSS-BF-BIC Continuous 10:2 & 100000 & 4 & 32.000000 & * \\
large & Continuous & PC-MAX-BF-LRT Continuous 10:2 & 10000 & 8 & * & 0.000000 \\
large & Continuous & PC-MAX-BF-LRT Continuous 10:2 & 20000 & 8 & * & 0.000000 \\
large & Continuous & PC-MAX-BF-LRT Continuous 10:2 & 50000 & 8 & * & 0.000000 \\
large & Continuous & PC-MAX-BF-LRT Continuous 10:2 & 100000 & 7 & * & 0.000000 \\
large & Mixed & BOSS-BF-BIC Mixed 10:2 & 10000 & 4 & 2.000000 & * \\
large & Mixed & BOSS-BF-BIC Mixed 10:2 & 20000 & 4 & 2.000000 & * \\
large & Mixed & BOSS-BF-BIC Mixed 10:2 & 50000 & 4 & 8.000000 & * \\
large & Mixed & BOSS-BF-BIC Mixed 10:2 & 100000 & 6 & 4.000000 & * \\
large & Mixed & PC-MAX-BF-LRT Mixed 10:2 & 10000 & 6 & * & 0.000100 \\
large & Mixed & PC-MAX-BF-LRT Mixed 10:2 & 20000 & 8 & * & 0.000100 \\
large & Mixed & PC-MAX-BF-LRT Mixed 10:2 & 50000 & 8 & * & 0.000100 \\
large & Mixed & PC-MAX-BF-LRT Mixed 10:2 & 100000 & 5 & * & 0.000000 \\
\bottomrule
\end{tabular}

\end{table}

\begin{table}[htbp]\centering\scriptsize

\caption{Optimal parameter settings by Sample Size (Large-P, Continuous Data)}
\label{tab:large_p_continuous_optimal_params}
\begin{tabular}{lllrrrr}
\toprule
Scale & Type & Label & sample\_size & trunc\_limit & penalty & alpha \\
\midrule
large & Continuous & BOSS-BF-BIC Continuous 50:4 & 500 & 3 & 1.000000 & * \\
large & Continuous & BOSS-BF-BIC Continuous 50:4 & 1000 & 3 & 1.000000 & * \\
large & Continuous & BOSS-BF-BIC Continuous 50:4 & 3000 & 3 & 2.000000 & * \\
large & Continuous & BOSS-BF-BIC Continuous 50:4 & 5000 & 3 & 2.000000 & * \\
large & Continuous & PC-MAX-BF-LRT Continuous 50:4 & 500 & 3 & * & 0.001000 \\
large & Continuous & PC-MAX-BF-LRT Continuous 50:4 & 1000 & 3 & * & 0.001000 \\
large & Continuous & PC-MAX-BF-LRT Continuous 50:4 & 3000 & 3 & * & 0.001000 \\
large & Continuous & PC-MAX-BF-LRT Continuous 50:4 & 5000 & 4 & * & 0.001000 \\
large & Continuous & BOSS-BF-BIC Continuous 100:4 & 500 & 3 & 1.000000 & * \\
large & Continuous & BOSS-BF-BIC Continuous 100:4 & 1000 & 3 & 1.000000 & * \\
large & Continuous & BOSS-BF-BIC Continuous 100:4 & 3000 & 3 & 2.000000 & * \\
large & Continuous & BOSS-BF-BIC Continuous 100:4 & 5000 & 3 & 2.000000 & * \\
large & Continuous & PC-MAX-BF-LRT Continuous 100:4 & 500 & 3 & * & 0.001000 \\
large & Continuous & PC-MAX-BF-LRT Continuous 100:4 & 1000 & 3 & * & 0.001000 \\
large & Continuous & PC-MAX-BF-LRT Continuous 100:4 & 3000 & 4 & * & 0.001000 \\
large & Continuous & PC-MAX-BF-LRT Continuous 100:4 & 5000 & 4 & * & 0.001000 \\
\bottomrule
\end{tabular}

\end{table}

\end{document}